%% file: main.tex
\documentclass[10pt,twocolumn,letterpaper]{article}

\usepackage{iccv}
\usepackage{times}
\usepackage{epsfig}
\usepackage{graphicx}
\usepackage{amsmath}
\usepackage{amssymb}

\input{macros}

\usepackage[breaklinks=true,bookmarks=false]{hyperref}

\iccvfinalcopy
\def\iccvPaperID{3381} %

\begin{document}

\title{Who's Waldo? Linking People Across Text and Images}

\author{
Claire Yuqing Cui$^{1*}$ \ 
Apoorv Khandelwal$^{1*}$ \ 
Yoav Artzi$^{1,2}$ \ 
Noah Snavely$^{1,2}$ \ 
Hadar Averbuch-Elor$^{1,2}$
\\[2mm]
$^1$Cornell University \
$^2$Cornell Tech
\\[1mm]
{\tt\small \{yc2296, ak2254, yoavartzi, snavely, hadarelor\}@cornell.edu}
}

\maketitle
\blfootnote{$^*$ Equal contribution}

\begin{abstract}
\input{00-abstract}
\end{abstract}

\input{01-intro}
\input{02-related}
\input{03-task}
\input{04-dataset}

\input{05-method}
\input{06-results}

\input{07-conclusion}

\medskip
\noindent \textbf{Acknowledgments.}
This work was supported by the National Science Foundation (IIS-2008313, CAREER-1750499), a Google Focused Award, the generosity of Eric \& Wendy Schmidt by recommendation of the Schmidt Futures program, and the Zuckerman STEM Leadership Program.

{\small
\bibliographystyle{ieee_fullname}
\bibliography{references}
}

\newpage
\input{supplementary/appendix}

\end{document}

%% file: macros.tex
\usepackage{xcolor}
\usepackage{booktabs}
\usepackage{multirow}
\usepackage{tabularx}
\usepackage{xspace}
\usepackage[accsupp]{axessibility}

\newcommand{\dataset}{\emph{Who's Waldo}\xspace}
\newcommand{\NAME}{\texttt{[NAME]}\xspace}

\definecolor{tomato}{RGB}{255,99,71}
\definecolor{gold}{RGB}{255,215,0}
\definecolor{mediumaquamarine}{RGB}{102,205,170}
\definecolor{blueviolet}{RGB}{138,43,226}
\definecolor{cornflowerblue}{RGB}{100,149,237}
\definecolor{lightpink}{RGB}{255,182,193}
\definecolor{orchid}{RGB}{218,112,214}

\newcommand\blfootnote[1]{%
  \begingroup
  \renewcommand\thefootnote{}\footnotetext{#1}%
  \endgroup
}

\newbox\jsavebox
\newcommand{\jsubfig}[2]{%
	\sbox\jsavebox{#1}%
	\parbox[t]{\wd\jsavebox}{\centering\usebox\jsavebox\\#2}%
	}

%% file: 00-abstract.tex
We present a task and benchmark dataset for \textbf{person-centric visual grounding}, the problem of linking between people named in a caption and people pictured in an image. In contrast to prior work in visual grounding, which is predominantly object-based, our new task masks out the names of people in captions in order to encourage methods trained on such image--caption pairs to focus on contextual cues, such as the rich interactions between multiple people, rather than learning associations between names and appearances. To facilitate this task, we introduce a new dataset, \textbf{\dataset}, mined automatically from image--caption data on Wikimedia Commons. We propose a Transformer-based method that outperforms several strong baselines on this task, and release our data to the research community to spur work on contextual models that consider both vision and language. Code and data are available at: \url{https://whoswaldo.github.io}

%% file: 01-intro.tex
\section{Introduction}

The correspondence between people observed in images and their mentions in text is informed by more than simply their identities and our knowledge of their appearances.
Consider the image and caption in Figure~\ref{fig:teaser}. We often see such image--caption pairs in newspapers and, as humans, are skilled at recovering associations between the people depicted in images and their references in captions, even if we're unfamiliar with the specific people mentioned. This ability requires complex visual reasoning skills. For the example in Figure~\ref{fig:teaser}, we must understand an underlying activity (``passing'') and determine who is passing the ball, who is being passed to, and which people in the image are not mentioned at all.

In this paper, we present a person-centric vision-and-language grounding task and benchmark.
The general problem of linking between textual descriptions and image regions is known as \emph{visual grounding}, and is a fundamental capability in visual semantic tasks with applications including image captioning~\cite{you2016image,lu2017knowing,anderson2018bottom}, visual question answering~\cite{fukui2016multimodal,Fukui:16bilinearpoolvqa,Hu:17compnetqa} and instruction following~\cite{Anderson:18r2r,Misra:17instructions,Chen19:touchdown}.
Our task and data depart from most existing works along two axes. 
First, our task abstracts over identity information, instead focusing specifically on the relations and properties specified in images and text. 
Second, rather than using data annotated
by crowd workers, we leverage captions originating from real-life data sources.

\input{figures/examples/teaser}

While visual grounding has traditionally centered around localizing objects based on referring expressions, we observe that inferring associations based on expressions in person-centric samples---\ie people's names---could lead to problematic  biases  (\eg with regards to gender).
Hence, we formulate the task to use captions that mask out people's names. This allows for an emphasized focus on the context---both in image and text---where the person appears, requiring models to understand complex asymmetric human interactions and expected behaviors. For instance, in the example in Figure~\ref{fig:teaser} we might expect a player to pass to someone on their own team.

To explore this problem, we create \dataset: a collection of nearly 300K images of people paired with textual descriptions and automatically annotated with alignments between mentions of people's names and their corresponding visual regions. 
\dataset is constructed from the massive public catalog of freely-licensed images and descriptions in Wikimedia Commons. We leverage this unique data source to automatically extract image--text correspondences for over 200K people. We also provide evaluation sets that are validated using Amazon Mechanical Turk and demonstrate that our annotation scheme is highly accurate.

To link people across text and images, we propose a Transformer-based model, building upon recent work on learning joint contextualized image--text representations. 
We use similarity measures in the joint embedding space between mentions of people and image regions depicting people to estimate these links. 
The contextualized Transformer-based representations are particularly suited to handle the masked names, by shifting the reasoning to surrounding contextual cues such as verbs indicating actions and adjectives describing visual qualities. 
Our results demonstrate that our model effectively distinguishes between different individuals in a wide variety of scenes that capture complex interactions, significantly improving over strong baselines.

%% file: figures/examples/teaser.tex
\begin{figure}
    \centering
    \includegraphics[width=0.915\columnwidth]{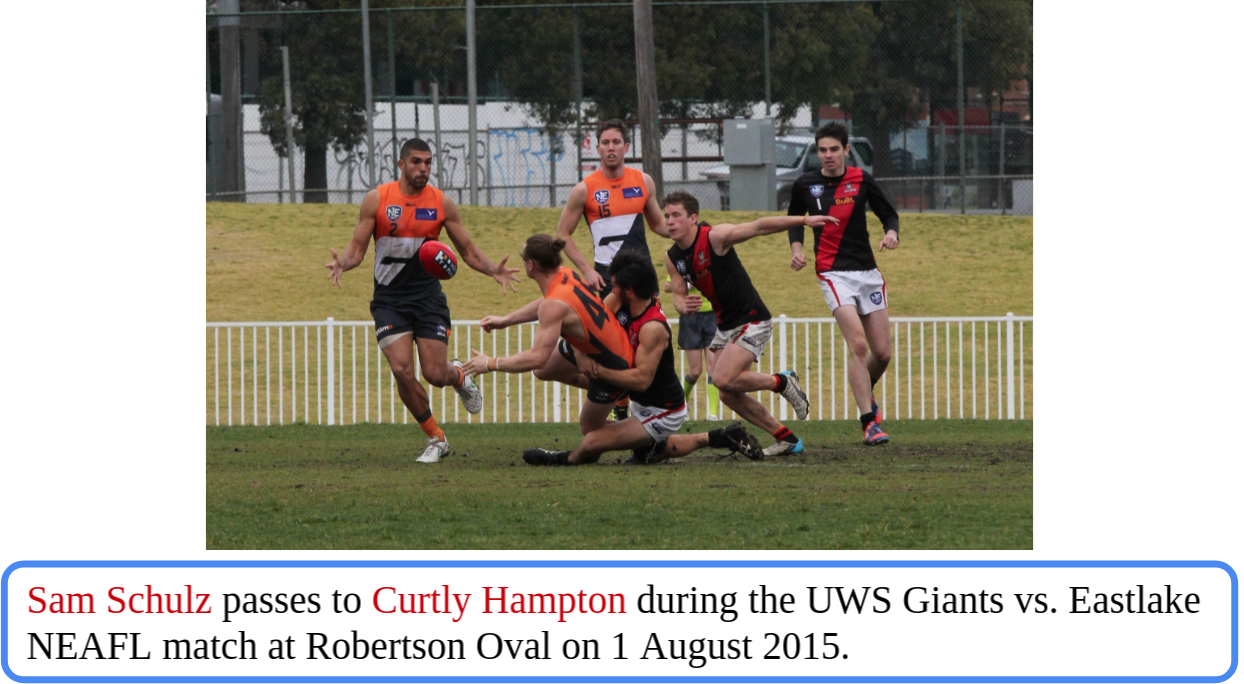}
    \caption{
By studying this picture and caption, we can use contextual cues to link between the people referred to in the text and their visual counterparts, even if we are unfamiliar with the specific individuals. This capability requires understanding of a broad set of interactions
(\eg ``passing'') and expected behaviors (\eg players pass to their teammates). We propose the task of \emph{person-centric visual grounding}, where we abstract over identity names (\eg masking out \emph{Sam Schulz} and \emph{Curtly Hampton} with \NAME tokens) to encourage algorithms to emulate such contextual reasoning.
  }
    \label{fig:teaser}
\end{figure}

%% file: 02-related.tex
\section{Related Work}
\label{sec:related}

\noindent \textbf{Visual Grounding.} 
The goal of visual grounding is to localize objects in an image given a textual description. Tasks are typically formulated to either recover correspondences between object region proposals and text, or compute attention maps over the whole image.
Referring expression comprehension (REC) is a common variant of this problem, where the goal is to identify an image region corresponding to a sentential description (\eg~\cite{rohrbach2016grounding,engilberge2018finding,deng2018visual,yu2018mattnet,liu2019learning}).
Sadhu et al.~\cite{sadhu2019zero} recently extended this task to a zero-shot setting that also considers expressions with unseen nouns. Qiao et al.~\cite{qiao2020referring} provide a comprehensive survey on REC. 

Still, this line of work has made limited use of descriptions of relationships between objects. The Flickr30K Entities dataset~\cite{plummer2015flickr30k} opened up new avenues for modeling such dependencies by including images, full captions, and ground-truth links between regions and phrases for nearly a hundred object categories. 
Several methods have since been proposed to visually ground objects from textual descriptions that describe multiple objects~\cite{wang2016structured,plummer2017phrase,liu2020phrase}. 
A weakly-supervised setting, which assumes that ground truth links between regions and phrases are not available, has also gained attention, 
with discriminative and contrastive objectives~\cite{xiao2017weakly,wang2020maf,gupta2020contrastive}, visual and linguistic consistencies~\cite{chen2018knowledge} and multilevel aggregation strategies~\cite{zhao2018weakly,akbari2019multi,datta2019align2ground} used to align the image and language spaces.

However, most existing tasks in visual grounding permit models to reason over referring expressions directly (allowing models to learn priors over different object categories). Our proposed task instead requires models to exclusively reason over context and interactions between objects, as the referring expressions (\ie names) are masked.

The creation of most datasets related to visual grounding involves a time-consuming, expensive annotation process that includes both (i) generating referring expressions or full textual descriptions for a given image, and (ii) annotating corresponding regions in the image (\eg~\cite{lin2014microsoft,kazemzadeh2014referitgame,plummer2015flickr30k,yu2016modeling}).
We instead construct \dataset through an automatic approach inspired by 
Conceptual Captions~\cite{sharma2018conceptual}. %
While that work uses alt-text image descriptions from HTML (which are noisy and must be aggressively filtered), we use raw descriptions obtained from captions in Wikimedia Commons.

\medskip \noindent \textbf{People-centric Tasks.} 
Person identification~\cite{brunelli1995person,layne2012towards,zheng2016person} is 
a task related to the one we propose, 
and is formulated as a comparison between reference and target images, aiming to determine whether these belong to the same identity. 
Our work instead focuses on learning a contextual correspondence between image regions and textual captions describing people and their depicted interactions. For ethical reasons (see Ethical Considerations in Section~\ref{sec:dataset}), our released dataset does not contain identity information, and thus cannot be easily modified to train such models. %

Another related people-centric task is 
to select a set of attributes that will generate a 
description for each person in an image that distinguishes that 
individual from others in that image~\cite{sadovnik2013s}. 
Finally, Aneja \etal detect out-of-context image and caption pairs, using a dataset collected from news and fact-checking websites. Their data (specifically, the subset of images capturing people) could  be used to augment ours.

\medskip \noindent \textbf{Task-agnostic Joint Image-Text Representations.} 
Recent advances have led to a surge of interest in task-agnostic joint visual and textual representations \cite{lu2019vilbert,tan2019lxmert,li2019visualbert,su2019vl,chen2020uniter,sun2019videobert,zhou2020unified,lu202012,li2020oscar,gan2020large}. Several works, such as LXMERT~\cite{tan2019lxmert} and ViLBERT~\cite{lu2019vilbert} learn these representations using two-stream transformers~\cite{vaswani2017attention} (one per modality). Others, including VisualBERT~\cite{li2019visualbert}, VL-BERT~\cite{su2019vl} and UNITER~\cite{chen2020uniter}, use a unified architecture. In our work, we leverage these task-agnostic features to learn to link between the individuals described in the text and their visual counterparts.

%% file: 03-task.tex
\section{Person-centric Visual Grounding}
\label{sec:task}
Given an image $I$ with $m \geq 1$ people detections and a corresponding caption $x_s$ referring to $n \geq 1$ people (with each person mentioned
one or more times), 
we wish to find 
a mapping from referred people to visual detections.

We expect to produce a partial, injective (one-to-one) mapping, %
since 
not all referred people will be pictured and
no two referred people should map to the same detection. We also find that this mapping is not necessarily surjective (onto), since the image may picture people who are not named in the caption and there could exist detections not mapped to by any referred people.

In-the-wild captions featuring people often refer to them by name. 
However, reasoning about visual grounding using actual names of people involves two challenges: the diversity of names creates significant data sparsity, and their surface form (\ie, the text itself) elicits strong biases, e.g., with regard to gender. 
We therefore abstract over the surface form of the names 
by replacing each name with the placeholder token \NAME. 
This encourages models to focus on the textual context of the names, including adjectives and adverbs that hint at the person's visual appearance and verbs that indicate the action they partake in. In other words, by masking names we seek models that do not memorize what specific people look like, or form stereotypical associations based on specific names, but must instead learn richer contextual cues.
As part of our dataset, we provide a mapping from referred people to their respective sets of referring \NAME tokens.

While visual grounding has traditionally centered around localization of objects (including unnamed people), we find that visual grounding in the context of named people (which we denote as \emph{person-centric}) presents additional opportunities. In object-centric visual grounding, referring expressions are not masked, allowing models to also learn by matching images and object classes, rather than entirely from context. Moreover, data for our task (\ie, captioned images of people) is readily available on the web and matches a realistic distribution more closely than object datasets, whose pairs are annotated by workers for the sole purpose of visual grounding tasks. %

\medskip \noindent \textbf{Evaluation.}
Given a mapping produced by an algorithm for an input example, we evaluate by computing accuracy against ground truth links of referred people and detections. This is unlike prior works that extract hundreds of candidate boxes and approximate correct matches using either intersection-over-union ratios or the pointing game, which requires the model to predict a single point per phrase.
We also enforce that the people in test images and captions do not appear during training.

%% file: 04-dataset.tex
\input{figures/examples/examples}

\vspace{-0.5cm}

\section{
The \dataset Dataset
\includegraphics[height=0.9cm]{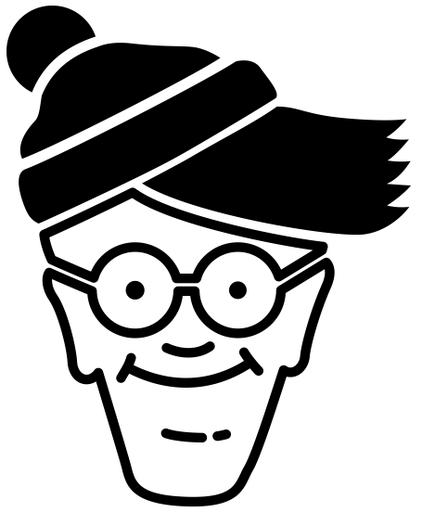}
}
\label{sec:dataset}

In this section, we describe \dataset \footnote{Icon created by Stefan Spieler from the Noun Project}, a new dataset with 270K image--caption pairs, derived from Wikimedia Commons.\footnote{\url{https://commons.wikimedia.org}}  We first describe the process of constructing and annotating this dataset, then present an analysis over dataset statistics. We show samples
from our dataset, along with their annotations, in Figure~\ref{fig:examples}.

\medskip \noindent \textbf{Data Collection.}
Under the broader ``People by name'' category in Wikimedia Commons there are 407K categories named after people, each with their own hierarchy of sub-categories. We refer to this set of people as \emph{Wikimedia identities}. 
We identified all sub-categories that are person-centric (\eg ``Barack Obama playing basketball'' or ``Sally Ride on Challenger in 1983'', rather than ``John F.\ Kennedy International Airport'') by tokenizing names, matching tokens with regular expressions, and tagging parts of speech. We then downloaded 3.5M images, 
collated duplicates, and retained references to the Wikimedia identities they originate from. We observe that images originating from an identity are very likely to depict that identity.

Many images on Wikimedia Commons are also associated with human-provided English captions that describe these images by naming the people present and detailing their settings and interactions. We collected these captions and pre-processed them by pattern matching with regular expressions to remove Wikimedia-specific text structures. We also removed phrases that are variants of ``photo by [photographer name]'', since photographers are often named in captions but are not pictured in images.

\medskip \noindent \textbf{Detecting People in Images and Captions.}
To detect 
bounding boxes for people in images, we used a Switchable Atrous Convolution model with a Cascade R-CNN and ResNet-50 backbone from MMDetection~\cite{qiao2020detectors, mmdetection} trained on COCO~\cite{lin2014microsoft}. We subsequently estimated 133 whole-body keypoints using a top-down DarkPose model from MMPose~\cite{zhang2020distribution, mmpose2020} (trained on COCO~\cite{lin2014microsoft} and finetuned on COCO-WholeBody~\cite{jin2020whole}).

We applied a pre-trained Punkt sentence tokenizer from NLTK~\cite{Kiss2006UnsupervisedMS, Bird2006NLTKTN} to all captions and performed named entity recognition on each sentence using FLAIR~\cite{Akbik2019FLAIRAE} to identify person names. %
We observe that people can be mentioned more than once in captions and without exact matches (\eg, as ``William'' and ``Bill'', or ``Barack'' and ``Obama''). Therefore, we used neural coreference resolution models from AllenNLP~\cite{Lee2017EndtoendNC, Gardner2018AllenNLPAD} to cluster multiple name entities as individual referred persons.

\medskip \noindent \textbf{Estimating Ground Truth Links.}
To produce supervision for our task, we automatically generated ground truth links from referred people in captions to detections of people in images. As we will describe, 
Wikimedia Commons provides reference faces for many referred people. As we can also generate face images for our image detections (via face alignment from estimated pose landmarks), we computed a similarity matrix using FaceNet embeddings~\cite{Schroff2015FaceNetAU, sandberg2017} between reference faces and detected faces. By finding a minimum weight bipartite matching~\cite{Kuhn1955TheHM} in this matrix and applying a threshold (set empirically to $0.46$), 
we recovered a partial mapping from referred people to detections.

We find reference faces for referred people as follows. First, we associate referred people with Wikimedia identities (via the prior coreference resolution step). We also find that many Wikimedia identities have \textit{primary} images on Wikimedia Commons, which prominently display their faces. We treat these as reference faces of referred people. However, not all referred people have such associations, so our ground truth links are a subset of all links.

\input{figures/statistics/statistics}

\medskip \noindent \textbf{Dataset Size and Splits.}
The above process yields
271,747 image--caption pairs. Figure~\ref{fig:statistics} summarizes the distributions of annotations and identities present in \dataset.

We split these into 179K training, 6.7K validation, and 6.7K test image--caption pairs. We generate the validation and test splits without overlapping identities from training and by ensuring that examples are challenging and correctly annotated. To do so, we first randomly select 16K identities and produce a validation and test superset from examples containing these identities (observing that additional identities  likely appear in these examples as well). We generate the training set from all remaining examples that do not contain \emph{any} identities in the superset. We then remove all (trivial) examples from the superset with exactly one person detection and one referred person. We manually validate this superset further 
as described below and divide the resulting examples into validation and test splits.

\input{figures/overview/overview}

\medskip \noindent \textbf{Validating Test Images with AMT.}
While our method approximates ground truth mappings, we want subsets for evaluation that only include correct ground truth links. To that end, we used Amazon Mechanical Turk (AMT) to remove test set examples with incorrect annotations. Given a ground truth link (\ie identity name and image crop of detected person), we defined the following yes/no AMT task: ``Does this [detection crop] contain [identity name]?''. For ease of comparison, we also provided workers with a reference image and a link to additional photos for that identity. We assign each ground truth link to two workers. Finally, we select all pairs for which both workers answered ``yes''.

We manually inspected 400 responses and---accounting for worker disagreement and error---estimate that our automatic technique was accurate for approximately $95.5\%$ of links in superset examples. However, after removing any examples for which either worker answered ``no'', we estimate that over $98.5\%$ of links in the retained examples are accurate. Please refer to the supplemental material for additional visualizations over our dataset and generated links.

\medskip \noindent \textbf{Ethical Considerations}
People-centric datasets pose ethical challenges. For example, ImageNet~\cite{deng2009imagenet} has been scrutinized based on issues inherited from the ``person'' category in WordNet~\cite{crawford2019excavating,yang2020towards}. Our task and dataset were created with careful attention to ethical questions, which we
encountered throughout our work. Access to our dataset will be provided for research purposes only and with restrictions on redistribution. Additionally, as we mask all names in captions, our dataset cannot be easily repurposed for unintended tasks, such as identification of people by name. Due to biases in our data source, we do not consider the data appropriate for developing %
non-research systems without further processing or augmentation. More details on distribution and intended uses are provided in a supplemental datasheet~\cite{gebru2018datasheets}.

%% file: figures/examples/examples.tex
\begin{figure*}
  \jsubfig{\includegraphics[height=4.85cm]{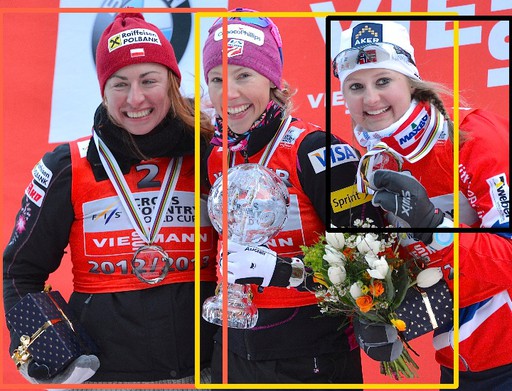}}{} 
        \hfill
\jsubfig{\includegraphics[height=4.85cm]{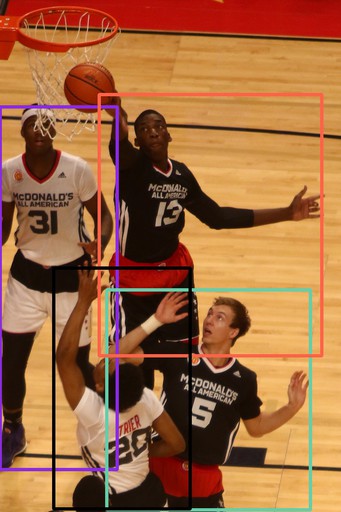}}{} 
\hfill
\jsubfig{\includegraphics[height=4.85cm]{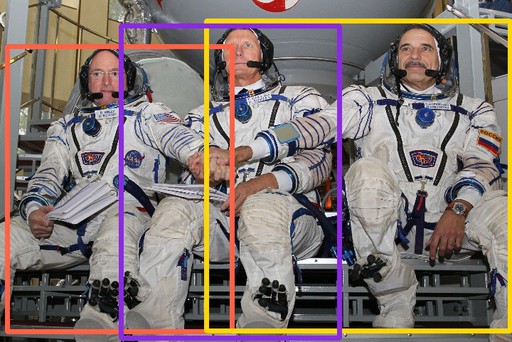}}{} 
 {\leftskip=0.1pt   \footnotesize{
  \textbf{\underline{Left}}: ``\colorbox{tomato}{\textbf{Justyna Kowalczyk}}, \colorbox{gold}{\textbf{Kikkan Randall}} and \textbf{Ingvild Flugstad Østberg} at the Royal Palace Sprint, part of the FIS World Cup 2012/2013, in Stockholm on March 20, 2013. \colorbox{gold}{\textbf{Kikkan Randall}} won the sprint cup.''
  \textbf{\underline{Center}}: ``\colorbox{tomato}{\textbf{Cheick Diallo}} blocks \textbf{Allonzo Trier} (\#20) in front of \colorbox{mediumaquamarine}{\textbf{Luke Kennard}} (\#5) and \colorbox{orchid}{\textbf{Carlon Bragg}} (\#31) in the 2015 McDonald's All-American Boys Game.''
  \textbf{\underline{Right}}: ``At the Gagarin Cosmonaut Training Center in Star City, Russia, Expedition 41/42 backup crew members \colorbox{tomato}{\textbf{Scott Kelly}} of NASA (left), \colorbox{orchid}{\textbf{Gennady Padalka}} of the Russian Federal Space Agency (Roscosmos, center) and \colorbox{gold}{\textbf{Mikhail Kornienko}} of Roscosmos (right) clasp hands as they pose for pictures in front of a Soyuz simulator at the start of final qualifications.''}} %
  \vspace{3pt}
  \caption{
  \textbf{Samples from \dataset}, showing detected named entities in bold and entities linked with image regions in unique colors, corresponding to the boxes on the images. Unmatched boxes and entities are colored in black.
  }\label{fig:examples}
\end{figure*}

%% file: figures/statistics/statistics.tex
\begin{figure} %
    \centering
\jsubfig{\includegraphics[height=2.83cm,,trim={0 0 0 1.0cm},clip]{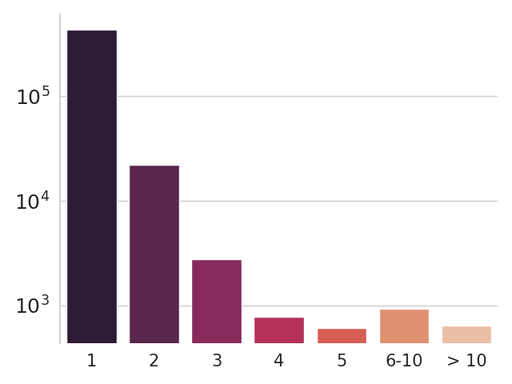}}{Mentions per Name} 
        \hfill
\jsubfig{\includegraphics[height=2.83cm,,trim={0 0 0 1.0cm},clip]{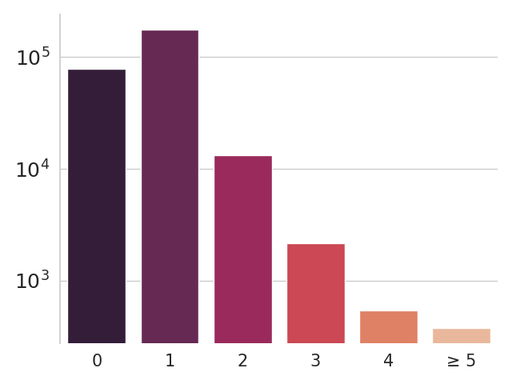}} {Links per Sample} \\
\jsubfig{\includegraphics[height=2.83cm,,trim={0 0 0 0.3cm},clip]{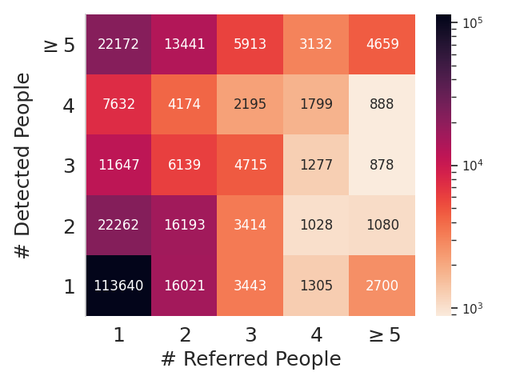}} {Distribution of Samples}
        \hfill
\jsubfig{\includegraphics[height=2.83cm,,trim={0 0 0 1.0cm},clip]{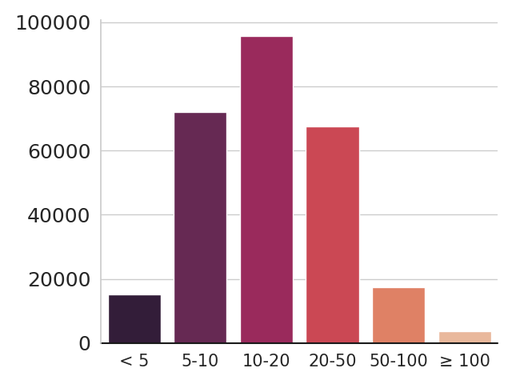}}{Caption Length (words)}
    \vspace{3pt}
    \caption{\textbf{\dataset statistics,} including the number of number of mentions (occurrences) for referred people in captions, ground truth box-name links per sample, distribution of samples and caption length (by words).
   }
    \label{fig:statistics}
\end{figure}

%% file: figures/overview/overview.tex
\begin{figure*} %
\jsubfig{
    \includegraphics[height=5.3cm]{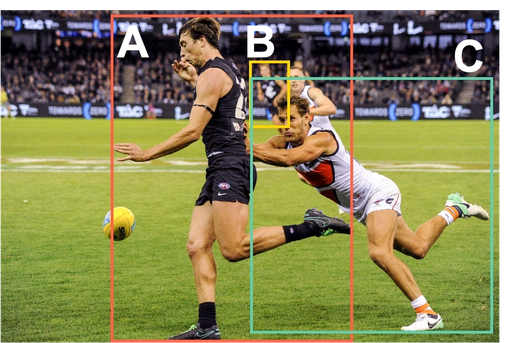}
}{
    \vspace{-200pt} \leftskip=0.1pt {\colorbox{tomato}{\textbf{Caleb Marchbank}} kicking away from \colorbox{mediumaquamarine}{\textbf{Matt de Boer}} during the AFL round twelve match between Carlton and Greater Western Sydney on 11 June 2017 at Etihad Stadium in Melbourne, Victoria.}
} 
\hfill
\jsubfig{
    \includegraphics[height=7.1cm]{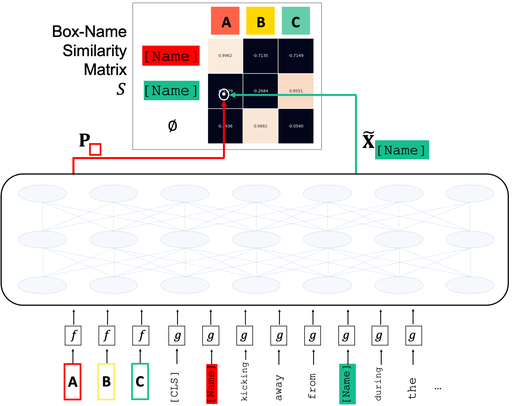}
}{}
\caption{\textbf{Overview of our approach.} Features are extracted from image regions and words and combined with a Transformer to learn similarities between people detected in the image (boxes A-C, colored in unique colors) and names mentioned in the caption (Caleb Marchbank and Matt de Boer in the example above). Correspondences are depicted by matching colors.}
\label{fig:overview}
\end{figure*}

%% file: 05-method.tex
\section{Method}\label{sec:method}

In this section we present an approach for linking people in text and images. We use a multi-layer Transformer~\cite{vaswani2017attention} to learn joint image--text representations such that referred people and their corresponding image regions will be highly similar, while those that do not correspond will be dissimilar.
For brevity, we refer to the $n$ names of referred people as \emph{names} and $m$ image regions of detected people as \emph{boxes}.

\subsection{Model}\label{sec:method:model}

Our method is based on the  recent UNITER Transformer model~\cite{chen2020uniter}. As shown in their work, their pretrained model can be leveraged for a wide variety of downstream vision-and-language tasks. In this section, we show how UNITER can be modified for our task and fine-tuned on our dataset. An overview of our approach is shown in Figure~\ref{fig:overview}.

We extract visual features for each person detection $p$ 
using a fully-convolutional variant of Faster R-CNN~\cite{anderson2018bottom}. Visual features are concatenated with encodings of their spatial coordinates,\footnote{Following \cite{chen2020uniter}, these are: $[x_1,y_1,x_2,y_2,w,h,w\times h]$.} yielding spatial-visual features $f(p)$.
We tokenize words into WordPieces~\cite{wu2016google}. In accordance with our task, names are symbolized by \NAME tokens. For each sub-word $w$, we extract features $g(w)$ that are composed of a token embedding and position embedding.

We feed these spatial-visual and textual features into a Transformer model that uses self-attention layers to learn a contextual representation and captures a more context-specific representation in upper-hidden layers~\cite{ethayarajh2019contextual}.
We denote the final hidden layer of spatial-visual features as $\mathbf{P}_k$ and of textual features as $\mathbf{X}_l$, where $\mathbf{P}_k,\mathbf{X}_l \in \mathbb{R}^{768}$. 

From these contextualized representations, we construct a box--name similarity matrix $S$ (top-right of Figure~\ref{fig:overview}). This matrix measures cosine similarities $S_{i,j}$ between the $i$-th name and the $j$-th box:
\begin{equation}
S_{i,j} = \frac{\mathbf{P}_j^T \tilde{\mathbf{X}}_i}{\|\mathbf{P}_j\|_2 \|\tilde{\mathbf{X}}_i\|_2},
\end{equation}
where $\tilde{\mathbf{X}}_i$ is an embedding averaged over all \NAME tokens for mentions of the $i$-th referred person in a caption. 

During inference, for each referred person, we select its corresponding detection 
as the most similar box in $S$. %

\subsection{Learning}\label{sec:method:learn}

To train our model, we propose 
the following loss terms that operate on the similarity matrix $S$: (1) box--name matching losses defined within and across images %
and (2) an unlinked box classification loss. 

\medskip 
\noindent
\textbf{Box--Name Matching Losses.} We define box--name matching losses within images (supervising estimated correspondences with ground truth links) and across images (using a discriminative objective over image--caption pairs).

We compute the estimated probability for a ground truth link $(i,j)$ over different boxes ($p = \mathrm{Softmax}(S_{i,:})_j$) and also over different names ($q = \mathrm{Softmax}(S_{:,j})_i$) in its corresponding image--caption pair. We minimize cross-entropy losses over these for all ground truth links $L$ in a batch:

\begin{equation}
\mathcal{L}_\mathsf{intra} = - \frac{1}{|L|} \sum_{l \in L} \left[ \log p^{(l)} + \log q^{(l)} \right]
\end{equation}
Because we would like to leverage additional images during training (\ie, those without ground truth links), we also compute a matching loss across images containing a single box and name (which are likely to represent the same person). We sample positive and negative box--name pairs. Negative pairs are generated by replacing the box with one from another image (and of a different person). We minimize a binary cross-entropy loss $\mathcal{L}_\mathsf{inter}$ over these pairs.

\input{figures/blur/blur}

\medskip \noindent \textbf{Unlinked Box Classification Loss.} 
As not all people depicted in an image are referred to in its caption, we augment $S$ with a constant null name $\tilde{\mathbf{X}}_{\varnothing}$. We formulate a binary cross-entropy classification loss over similarities between boxes and $\tilde{\mathbf{X}}_{\varnothing}$. We process  these similarities $S_{i=\varnothing,j}$ through a sigmoid function to obtain normalized values. Boxes linked to names are considered negative matches (\ie, these should yield low similarities with $\tilde{\mathbf{X}}_{\varnothing}$). 

We cannot assume all other boxes are positive matches (\ie, should yield high similarities with $\tilde{\mathbf{X}}_{\varnothing}$) as we are only provided with partial ground truth correspondences 
from the algorithm in Section \ref{sec:dataset}. Instead, we select unlinked boxes that are (1) \emph{insignificant} compared to other boxes in the image and (2) \emph{blurry}. Both are measured using the a detected person's face (computed from whole body landmarks): a face image $f$ is considered insignificant if $\mathrm{Area}(f) < 0.6 \cdot \mathrm{Area}(f_\mathrm{largest})$ and blurry if $\mathrm{Var}(\Delta(f)) < 50$~\cite{pertuz2013analysis}, where $f_\mathrm{largest}$ is the largest face in the image and $\Delta$ is the Laplace operator. Figure~\ref{fig:blur} shows several images from our dataset with unlinked boxes in red. We minimize a binary cross-entropy loss $\mathcal{L}_{\varnothing}$ over images containing such positive and negative matches.  

This loss, in addition to providing us a means of directly estimating whether or not a given box is referred to in a caption, also implicitly encourages the contextualized representations of insignificant and blurry faces to be distinguishable from others. As we show in our results, this improves the accuracy of identifying referred people, allowing the model to focus on more relevant boxes.

%% file: figures/blur/blur.tex
\begin{figure} %
    \centering
\jsubfig{\includegraphics[height=3.05cm]{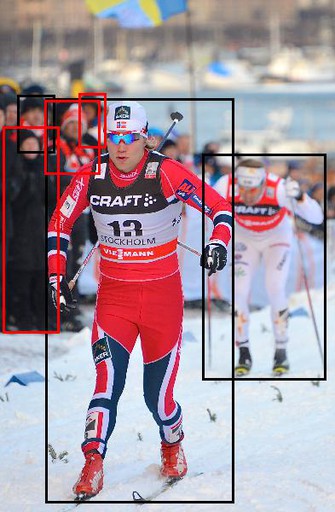}}{} 
        \hfill
\jsubfig{\includegraphics[height=3.05cm]{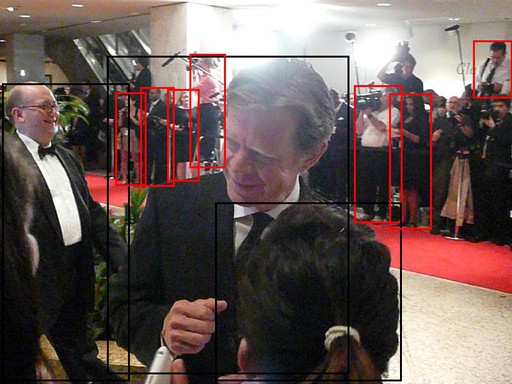}}{}
\hfill
\jsubfig{\includegraphics[height=3.05cm]{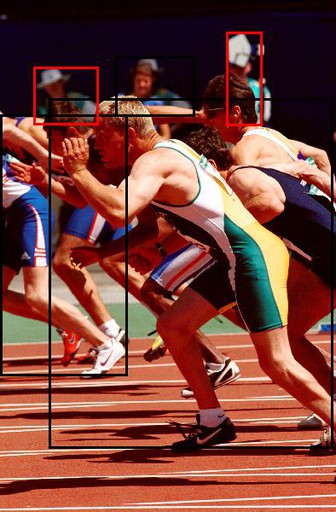}}{}
    \caption{\textbf{Selecting unlinked boxes.} We select small and blurry boxes (colored in red) for our proposed classification loss, encouraging the model to focus on larger (and less blurry) people.    
   }
    \label{fig:blur}
\end{figure}

%% file: 06-results.tex
\section{Results and Evaluation}\label{sec:results}

We compare our model
to other visual grounding methods trained on a variety of datasets. We study four key questions: How well do previous methods for visual grounding perform on our proposed task? To what extent is our model reasoning over complex multimodal signals? What is the impact of our design choices? And, what has our model learned?  
We also present qualitative results (Figure~\ref{fig:results_paper} and supplemental material), which  highlight the complexity and unique challenges of our proposed task.

\input{tables/evaluation}

\input{figures/examples/results2}

\subsection{Comparison to Prior Work}

We evaluate several recent visual grounding models on the \dataset test set: a weakly-supervised framework by Gupta \etal~\cite{gupta2020contrastive}, a supervised neural chain conditional random field that captures entity dependencies (SL-CCRF)~\cite{liu2020phrase}, and a supervised network that combines attention from separate modules (MAttNet)~\cite{yu2018mattnet}.
We also evaluate UNITER~\cite{chen2020uniter}, a pretrained multi-task vision-and-language framework, which our model is based on. %

Table~\ref{tab:evaluation} shows test set accuracies for our approach and for existing methods trained on different datasets. We report 95\% binomial proportion confidence intervals (Wilson score intervals) with these accuracies.
For existing models, we vary how names are provided during inference because these models are not automatically compatible with our placeholder \NAME token: (a) unmodified \textbf{full names}, (b) \textbf{random} popular names, %
or (c) a \textbf{constant} ``person'' string---\eg, ``Harry met Sally'' is modified to ``person met person''. %

We also evaluate several heuristics that illustrate the challenges and biases in our data (Table~\ref{tab:evaluation}), such as a potential left-to-right bias for named individuals. In particular, we order the names in the caption from left to right, and pair them with detections sorted by (a) decreasing area (Big$\rightarrow$Small), (b) left-to-right upper-left coordinates (L$\rightarrow$R (All)), or (c) left-to-right upper-left coordinates with only the largest $d$ detections (L$\rightarrow$R (Largest)).
We set $d = max(m, n)$ for $m$ detections and $n$ names. We also compare to random guessing.
We observe that these heuristics yield non-trivial, and even strong, performances. This could be because realistic captions tend to follow a left-to-right ordering (especially for posed people---but see Figure~\ref{fig:results_paper} for counterexamples) and filtering by detection size can remove unreferred people.
However, even the strongest heuristic leaves much room for improvement. 
These heuristics are also useful to frame the performance of pretrained visual grounding models. Supervised models (SL-CCRF and MAttNet) perform similarly to Big$\rightarrow$Small, illustrating that these models may be utilizing size-related cues---especially MAttNet, which \emph{only} processes names and not full sentences. We show qualitative results for all baselines in the supplemental material.

\subsection{Ablation study}

\input{tables/ablations}

Table~\ref{tab:ablations} shows ablation results. 
We train models using only a subset of features by ablating (i) visual features: set instead to a fixed representation, averaged over $1000$ random detections; (ii) spatial features: fixed at image center coordinates; (iii) textual embeddings: with all words masked out, retaining only position features and special \NAME tokens; and (iv) textual and visual embeddings: retaining only spatial features.
The impact of each input modality is significant, with performance dropping by $5.5\%$ for (ii) and $12.2\%$ for (iii). While these ablations significantly limit the information available for this task, our model performs much better than random guessing in all cases, suggesting that it learns some data biases. Both (i) and (iii) are capable of learning a left-to-right association. Indeed, their correct matches significantly overlap with those of the ``L$\rightarrow$R (Largest)'' heuristics, by $81.7\%$ for (i) and $82.4\%$ for (iii). Finally, from (iv) we infer that spatial features alone are not enough for learning such similarities.

We also quantify the importance of each proposed objective. Training without estimated correspondences (\ie, $\mathcal{L}_\mathsf{intra}$) yields the most significant drop in performance, resulting in nearly random guessing. This illustrates the importance of supervised data for our task. Ablating the other losses ($\mathcal{L}_\mathsf{inter}$ and $\mathcal{L}_\mathsf{\varnothing}$) degrades performance by only $1.7\%$. The relatively small impact of $\mathcal{L}_\mathsf{inter}$ highlights the importance of having many samples that capture interactions between multiple people, rather than samples with just one detection and referred person.

We also report the performance obtained by training our full model 
from scratch, without UNITER's pretrained weights. This leads to a large drop in performance ($>13\%$).

\subsection{Analysis of results}

We analyze the performance of our model over different test subsets to better understand what the model is learning. 
We observe that our model is more robust to a larger number of faces compared to L$\rightarrow$R (Largest). For instance, in the case of only one referred person in an image, our model retains high performance over an increasing number of faces, while the heuristic drops by almost $20\%$ (from an accuracy of $84.5\%$ for two detected people down to an accuracy of $67.6\%$ for four or more detected people). We further demonstrate this breakdown in the supplemental material.
Another subset we consider is an \emph{interactive} subset of test samples (\ie those with at least two detections and referred people and a verb in their caption). 
This potentially more challenging subset constitutes nearly one-third of our test set.
Our model's performance drops to $52.1\%$, while the baseline performance drops to $45.0\%$.

We also analyzed whether having multiple mentions of a person's name affects performance. Approximately $3\%$ of referred people in the test set are mentioned multiple times in the caption. For these identities, our model has a modest improvement of $2.1\%$ if provided additional mentions. This illustrates that our model can leverage additional information from co-occurrences in a caption to some extent.
Finally, we also analyze the performance of our method over several categories of identity occupations in the supplemental material, as we observe that these correlate well with different situations captured by our dataset.

\subsection{Limitations}

The complexity of certain interactions, such as in sports games where players compete closely together, poses challenges, not only to our model, but also to the person detector and our method for estimating ground truth links. Figure~\ref{fig:limitations} (left) demonstrates an example of a basketball game where the bodies of players overlap, thus some are not detected. The example on the right illustrates a failure of our model, where the interaction ``blocks'' is not correctly interpreted.

Further, some captions are insufficient to produce meaningful links. For example, in Figure~\ref{fig:limitations} (center), after replacing the names ``Joe Jonas'' and ``Demi Lovato'' with \NAME, it is impossible to tell which performer each corresponds to. Hence our model resorts to a simple left-to-right heuristic.

\input{figures/examples/limitations}

%% file: tables/evaluation.tex
\begin{table}[t]
  \centering
  \setlength{\tabcolsep}{3.2pt}
  \def\arraystretch{0.95}
  \begin{tabularx}{1.0\columnwidth}{lllcccccc}
    \toprule
    Method                                   & Training Data      &   & Accuracy    \\
    \midrule
    \textbf{Full Names} \\
    Gupta et al.~\cite{gupta2020contrastive} & COCO               &   & 36.9 $\pm$ 1.04          \\
    Gupta et al.~\cite{gupta2020contrastive} & Flickr30K Entities &   & 39.3 $\pm$ 1.05          \\
    SL-CCRF~\cite{liu2020phrase}             & Flickr30K Entities &   &  43.5 $\pm$ 1.06         \\
    MAttNet~\cite{yu2018mattnet}             & RefCOCOg           &   &  43.6 $\pm$ 1.06         \\
    UNITER~\cite{chen2020uniter}             & Multiple~\cite{lin2014microsoft,krishna2017visual,ordonez2011im2text,sharma2018conceptual}  &   &  36.3 $\pm$ 1.03            \\
    \midrule
    \textbf{Random} \\
    Gupta et al.~\cite{gupta2020contrastive} & COCO               &   &  39.3 $\pm$ 1.05         \\
    Gupta et al.~\cite{gupta2020contrastive} & Flickr30K Entities &   &  41.1 $\pm$ 1.06          \\
    SL-CCRF~\cite{liu2020phrase}             & Flickr30K Entities &   & 44.1 $\pm$ 1.07          \\
    MAttNet~\cite{yu2018mattnet}             & RefCOCOg           &   &  44.0 $\pm$ 1.07          \\
    UNITER~\cite{chen2020uniter}             & Multiple~\cite{lin2014microsoft,krishna2017visual,ordonez2011im2text,sharma2018conceptual}  &   &  38.4 $\pm$ 1.04          \\
    \midrule
    \textbf{Constant} \\
    Gupta et al.~\cite{gupta2020contrastive} & COCO               &   &  35.6 $\pm$ 1.03          \\
    Gupta et al.~\cite{gupta2020contrastive} & Flickr30K Entities &   &  38.2 $\pm$ 1.04          \\
    SL-CCRF~\cite{liu2020phrase}             & Flickr30K Entities &   &  46.4 $\pm$ 1.07          \\
    MAttNet~\cite{yu2018mattnet}             & RefCOCOg           &   &  24.1 $\pm$ 0.92           \\
    UNITER~\cite{chen2020uniter}             & Multiple~\cite{lin2014microsoft,krishna2017visual,ordonez2011im2text,sharma2018conceptual}  &   & 34.2 $\pm$ 1.02          \\
    \midrule
    Random                             & --                 &   & 30.9 $\pm$ 0.99          \\
    Big$\rightarrow$Small                       & --                 &   &  48.2 $\pm$ 1.07           \\
    L$\rightarrow$R (All)                            & --                 &   &  38.4 $\pm$ 1.04            \\
    L$\rightarrow$R (Largest)                           & --                 &   &  57.7 $\pm$ 1.06            \\
    \midrule
    Ours                                     & \dataset              && \textbf{63.5} $\boldsymbol{\pm}$ \textbf{1.03} \\
    \bottomrule
  \end{tabularx}
  \vspace{3pt}
  \caption{Evaluation on the \dataset test set. We compare against prior grounding methods using multiple configurations, varying according to how names are processed. We also compare to several simple baselines, detailed in the text.}
\label{tab:evaluation}
\end{table}

%% file: figures/examples/results2.tex
\begin{figure*} %
\centering

\jsubfig{\includegraphics[height=4.2cm,trim={0 1.0cm 0 2.0cm},clip]{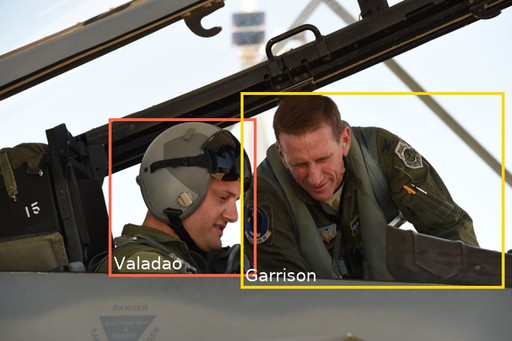}}{\vspace{-8pt}\begin{flushleft}
U.S. Air Force Colonel \colorbox{gold}{\textbf{Clay Garrison}} goes over some final instructions with U.S. Congressman \colorbox{tomato}{\textbf{David Valadao}} prior to take-off from the Fresno Air National Guard Base May 27, 2015.\end{flushleft}} \hfill
\jsubfig{\includegraphics[height=4.2cm]{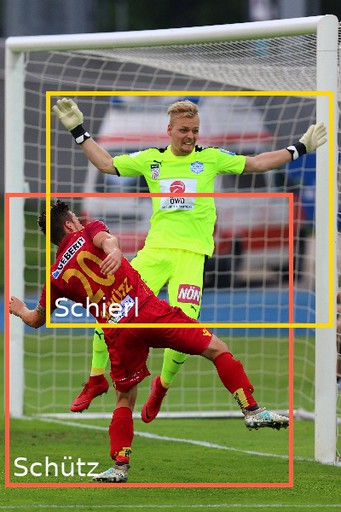}}{\vspace{-8pt}{\begin{flushleft} \colorbox{tomato}{\textbf{Daniel Schütz}} (\#20) behind goalkeeper \colorbox{gold}{\textbf{Domenik Schierl}}.\end{flushleft}}} \hfill
\jsubfig{\includegraphics[height=4.2cm]{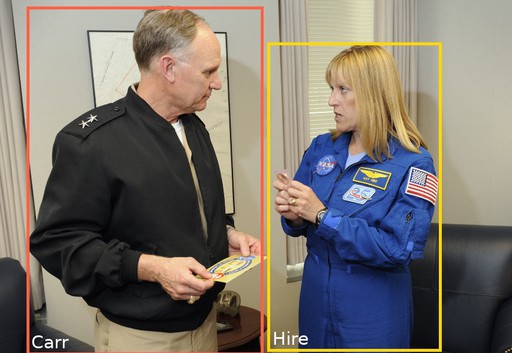}}{\vspace{-8pt}\begin{flushleft} \textbf{Kathryn Hire}, an astronaut and Navy reserve component Sailor assigned to the Office of Naval Research, presents items she took to space to Rear Adm. \colorbox{tomato}{\textbf{Nevin Carr}}.%
\end{flushleft}}

\vspace{-7pt}
\caption{\textbf{Box--name correspondences predicted by our model.} We show predicted entities on top of the their associated box (in white). Ground truth links are denoted by matching colors. Please refer to the supplemental material for additional qualitative results.}
\label{fig:results_paper}
\end{figure*}

%% file: tables/ablations.tex
\begin{table}[t]
\centering
\setlength{\tabcolsep}{3.1pt}
\def\arraystretch{0.95}
\begin{tabularx}{0.655\columnwidth}{lccc}
\toprule
  Method  && Accuracy        \\ \midrule
  \textbf{Input features} \\
 \quad w/o visual features && 55.4 $\pm$ 1.07  \\
 \quad w/o spatial features &&  58.0 $\pm$ 1.06 \\
 \quad w/o textual features && 51.3 $\pm$ 1.07 \\
  \quad spatial features only && 31.2 $\pm$ 0.99 \\
 \midrule
 \textbf{Learning} \\
 \quad w/o $\mathcal{L}_\mathsf{intra}$ && 31.4 $\pm$ 1.00 \\
  \quad  w/o $\mathcal{L}_\mathsf{inter}$ &&  61.9 $\pm$ 1.04 \\
    \quad  w/o $\mathcal{L}_{\varnothing}$ && 61.7 $\pm$ 1.04 \\
 \quad w/o pretraining &&  50.2 $\pm$ 1.07 \\
\bottomrule
\end{tabularx}
\vspace{3pt}
\caption{Ablation study, evaluating the effect of using different input features, loss terms and the impact using a pretrained model. }\label{tab:ablations}
\end{table}

%% file: figures/examples/limitations.tex
\begin{figure}
  \jsubfig{\includegraphics[height=2.75cm]{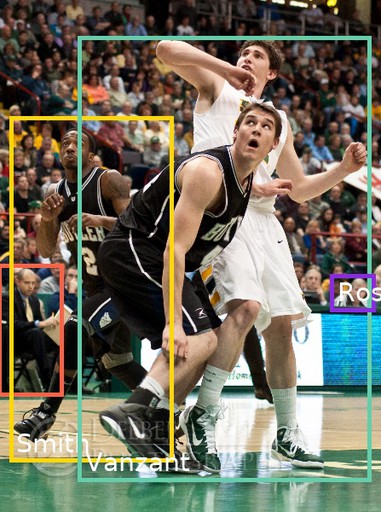}}{} 
  \hfill
\jsubfig{\includegraphics[height=2.75cm,trim={3cm 0 3.5cm 0},clip]{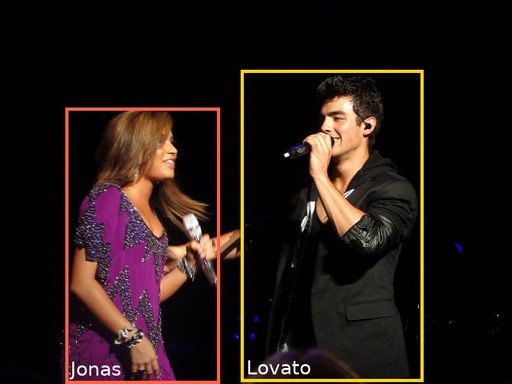}}{} 
\hfill
\jsubfig{\includegraphics[height=2.75cm,,trim={0.5cm 0 1cm 0},clip]{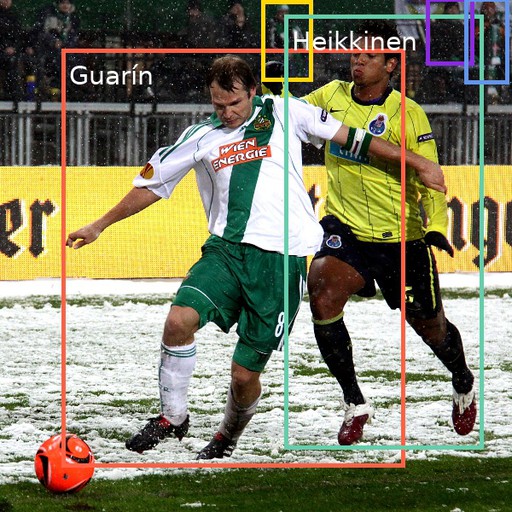}}{} 
\newline
 {\leftskip=0.1pt   \footnotesize{
  \textbf{\underline{Left}}: ``Butler's \colorbox{mediumaquamarine}{\textbf{Andrew Smith}} and Siena's \textbf{Ryan Rossiter} both try to anticipate the rebound, as Butler's \textbf{Shawn Vanzant} closes in from behind.''
  \newline
  \textbf{\underline{Center}}: ``\colorbox{gold}{\textbf{Joe Jonas}} and \colorbox{tomato}{\textbf{Demi Lovato}} performing in the Jonas Brothers Live In Concert.''}
  \newline
  \textbf{\underline{Right}}:``\colorbox{tomato}{\textbf{Markus Heikkinen}} blocks \colorbox{mediumaquamarine}{\textbf{Freddy Guarín}}.''}
  \vspace{3pt}
  \caption{
  \textbf{Examples our model predicted incorrectly}, showing detected named entities in bold and entities linked with image regions in unique colors, corresponding to the boxes on the images. %
  }\label{fig:limitations}
\end{figure}

%% file: 07-conclusion.tex
\section{Conclusion}\label{sec:conclusion}

We present a task, dataset, and method for linking people across images and text. 
By masking out names of people, we force methods to not memorize the appearance of specific individuals, but to understand contextual cues and interactions between multiple people.
Our approach shows encouraging performance on this task, but also indicates that the underlying task is very challenging and, as such, there is ample room for improvement via future methods that leverage our data. In particular, the performance of all methods drops given examples involving actions (as indicated by captions with verbs) and as the number of people referred to in a caption grows, indicating unresolved challenges in scaling to complex scenarios.

%% file: supplementary/appendix.tex
\appendix
\input{supplementary/01-dataset-details}
\input{supplementary/02-implementation-details}
\input{supplementary/03-baselines}
\input{supplementary/04-additional-results}

%% file: supplementary/01-dataset-details.tex
\section{Dataset Visualizations and Details}
Please refer to the accompanying \url{dataset_examples.html} for samples from our \dataset dataset. 

Our dataset has 215K ground truth links in total (for 193K images). Our dataset originates from over 400K Wikimedia identities and has ground truth links for 93K.

All images originate from Wikimedia Commons under free licenses. We group the licenses by freedom\footnote{\url{https://en.wikipedia.org/wiki/Free_license\#By_freedom}} as in Table~\ref{tab:licenses}.

We include a word cloud of the verbs present in our dataset in Figure~\ref{fig:verbs}.

\input{supplementary/figures/statistics/licenses}
\input{supplementary/tables/licenses}
\input{supplementary/figures/statistics/verbs}

\input{supplementary/figures/statistics/identity_occupations}
\input{supplementary/figures/statistics/poses}

\input{supplementary/figures/statistics/image_res}
\input{supplementary/figures/statistics/detection_res}

Our dataset contains images for at least 263K male and 70K female Wikimedia identities (these are identities we have labels for). We acknowledge this imbalance in ratio and attribute this to existing biases in our data source. However, our dataset is large enough that one could sample a more balanced subset. Our dataset does present diversity in the occupations of identities, as can be seen in Figure~\ref{fig:occupations}.

We show distributions of image resolutions in Figure~\ref{fig:image_res} and sizes of detection boxes (relative to image sizes) in Figure~\ref{fig:detection_res}. We show a distribution of head poses in our dataset in Figure~\ref{fig:poses}.

%% file: supplementary/figures/statistics/licenses.tex
\begin{figure}
    \centering
    \includegraphics[width=0.27\textwidth]{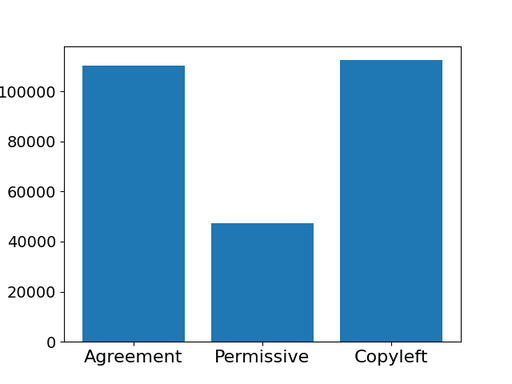}
    \caption{Distribution of copyright license groups for images in \dataset.}
    \label{fig:licenses}
\end{figure}

%% file: supplementary/tables/licenses.tex
\begin{table}
    \begin{tabularx}{\columnwidth}{cl}
	\toprule 
	Agreement & Copyrighted free use, No restrictions, CC0 \\
	& Public Domain, Public domain, WTFPL \\
	\midrule 
	Permissive & MIT, BSD, CC BY 1.0, \\
	& CC BY 2.0, CC BY 2.5, CC BY 3.0, \\
	& CC BY 4.0, Attribution, OGDL, \\
	& Licence Ouverte, KOGL Type 1, \\
	& OGL-C 2.0, OSPL, GODL-India, Beerware \\
	\midrule 
	Copyleft & GPL, GPLv2, GPLv3, LGPL, CC SA 1.0, \\
	& CC BY-SA 2.0, CC BY-SA 2.5, \\
	& CC BY-SA 3.0, CC BY-SA 4.0, \\
	& Nagi BY SA, GFDL 1.1, GFDL 1.2, \\
	& GFDL 1.3, GFDL, ODbL, OGL, OGL 2, \\
	& OGL 3, FAL, CeCILL \\
	\bottomrule
	\end{tabularx}
	\caption{Free licenses for images in our dataset (organized by freedom).}
    \label{tab:licenses}
\end{table}

%% file: supplementary/figures/statistics/verbs.tex
\begin{figure}
    \centering
    \includegraphics[width=0.77\columnwidth]{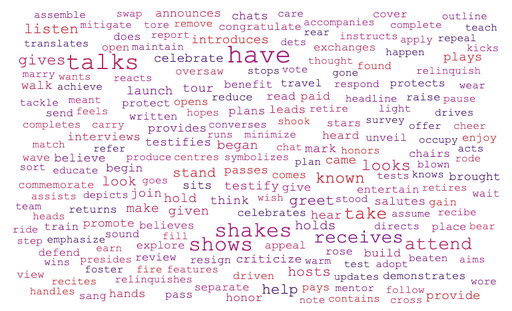}
    \caption{Visualization of verbs appearing in our dataset's captions. Larger font size correspond to verbs that appear more frequently in the dataset.}
    \label{fig:verbs}
\end{figure}

%% file: supplementary/figures/statistics/identity_occupations.tex
\begin{figure*}[!ht]
    \centering
    \includegraphics[width=\textwidth]{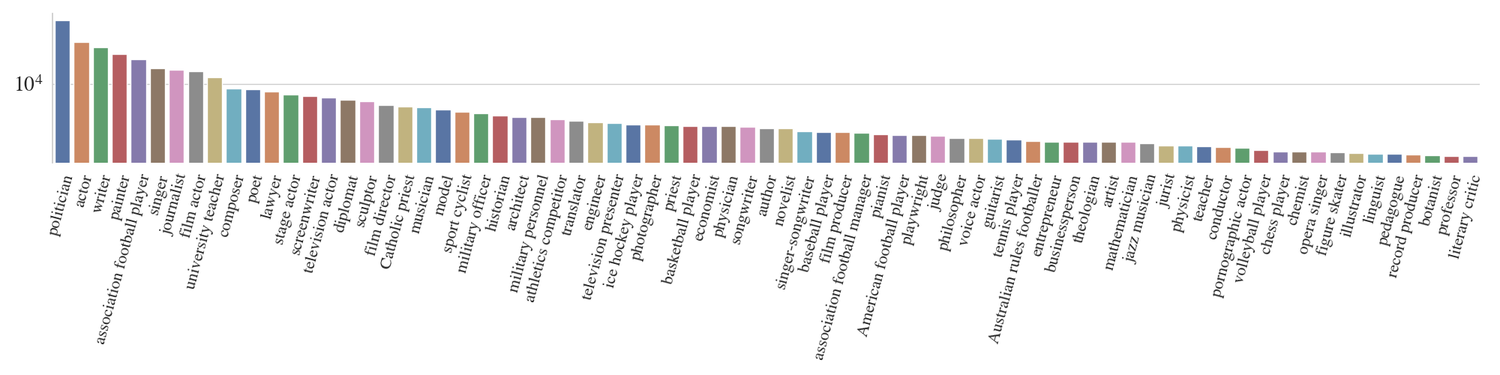}
    \caption{Distribution of occupations for Wikimedia identities in our dataset.}
    \label{fig:occupations}
\end{figure*}

%% file: supplementary/figures/statistics/poses.tex
\begin{figure*}[ht!]
\jsubfig{\includegraphics[height=4.4cm]{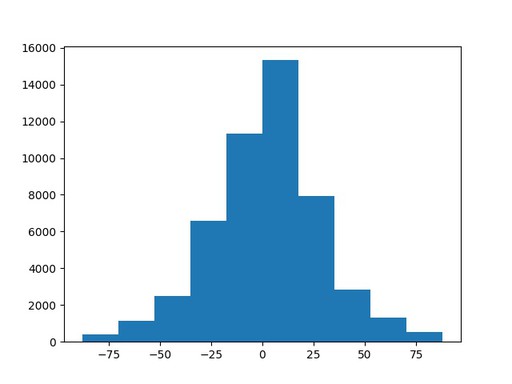}}{\vspace{-4pt}Yaw}
\jsubfig{\includegraphics[height=4.4cm]{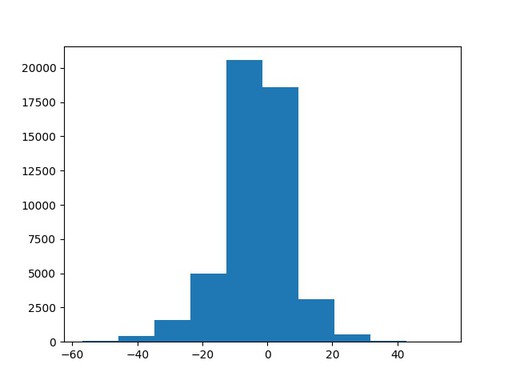}}{\vspace{-4pt}Pitch}
\jsubfig{\includegraphics[height=4.4cm]{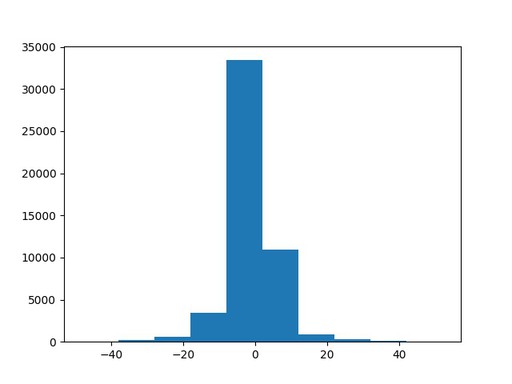}}{\vspace{-4pt}Roll}
\vspace{3pt}
\caption{Distribution of faces by degree of pose (head orientation) from a random subset of 50,000 detections. Note: yaw is the primary indicator of diversity in pose, as pitch and roll are limited by physical constraints for head rotation.}
\label{fig:poses}
\end{figure*}

%% file: supplementary/figures/statistics/image_res.tex
\begin{figure}[ht!] %
    \centering
\jsubfig{\includegraphics[height=2.83cm,,trim={0 0 0 1.0cm},clip]{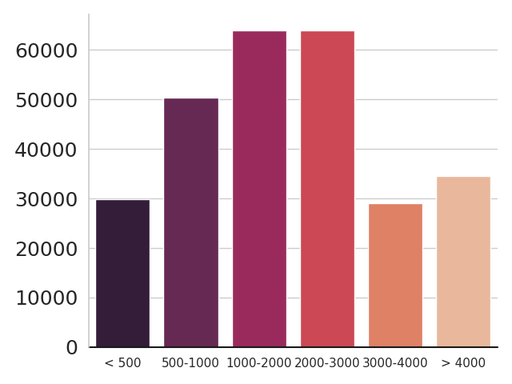}}{Width} 
        \hfill
\jsubfig{\includegraphics[height=2.83cm,,trim={0 0 0 1.0cm},clip]{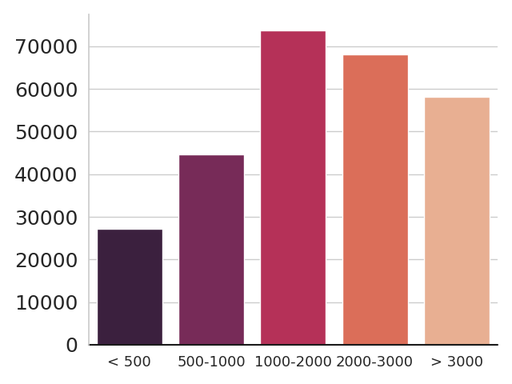}} {Height} \\
    \vspace{3pt}
    \caption{Distribution of image resolutions in our dataset.}
    \label{fig:image_res}
\end{figure}

%% file: supplementary/figures/statistics/detection_res.tex
\begin{figure}[ht!] %
    \centering
\jsubfig{\includegraphics[height=2.83cm,,trim={0 0 0 1.0cm},clip]{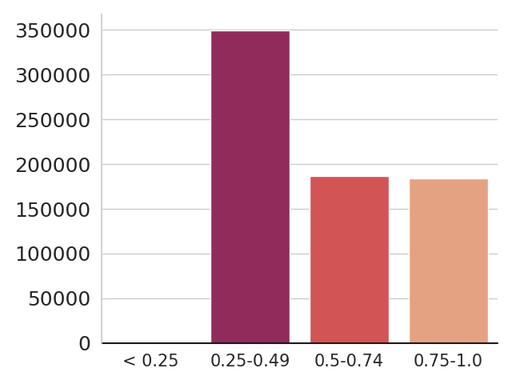}}{Width} 
        \hfill
\jsubfig{\includegraphics[height=2.83cm,,trim={0 0 0 1.0cm},clip]{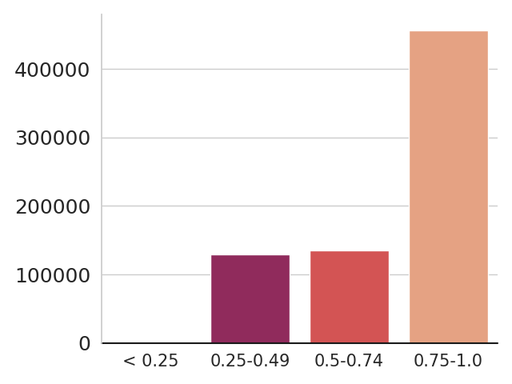}} {Height} \\
    \vspace{3pt}
    \caption{Distribution of detection sizes in our dataset.}
    \label{fig:detection_res}
\end{figure}

%% file: supplementary/02-implementation-details.tex
\section{Implementation Details}

\subsection{Dataset Cleaning}

We filter our data by removing all examples in which there are no people detected in an image or no people referred to in captions. We remove examples with captions that don't contain verbs or words other than names and stop words (\ie insubstantial captions). We further cleanse this data by removing images taken before 1990 (according to metadata) as we found this was a significant source of noise. We also found the presence of ``cropped'' versions of images that can be detected directly from file names containing the word ``cropped'', which usually only picture one person but have captions implying the presence of multiple, and also removed these.

\input{supplementary/figures/gt/gt}

\subsection{Training details}
We download the pretrained UNITER~\cite{chen2020uniter} model (UNITER-base). We use the ``bert-base-cased'' vocabulary from pytorch-transformers and add the [\texttt{NAME}] token. Following their implementation\footnote{\url{https://github.com/ChenRocks/UNITER}}, we define two training tasks that use two non-overlapping subsets of our dataset: ${(1,1)}$, containing images with exactly one referred person and one person box detected in the image, and ${(m,n)}$, containing all other images (\ie more than one referred person \emph{or} more than one box). 

The first task, denoted as \textbf{Task-1-1}, trains on the $(1,1)$ subset using the $\mathcal{L}_\mathsf{inter}$ objective, with $0.5$ probability of negative sampled image-caption pairs. The second task, denoted as \textbf{Task-M-N}, trains on the $(m,n)$ subset using the $\mathcal{L}_\mathsf{intra}$ and $\mathcal{L}_\mathsf{\varnothing}$ objectives. Furthermore, regarding $\mathcal{L}_\mathsf{intra}$, we note that this loss a sum over two cross-entropy losses, one over different boxes in the image and the other over different names in the caption.
Task-1-1 and Task-M-N are trained using a $1:2$ ratio.

We train 50,000 steps, validating performance over the validation set every 500 steps, with batch size of 1024. The max caption length we consider is 60 tokens, and the number of bounding boxes we consider is between 1 and 100, inclusive. Image-caption pairs not within these boundaries are filtered out during training. 
We use a learning rate of $5e-5$, weight decay of $0.01$ and dropout $0.1$, consistent with the default UNITER parameters (all other parameters are also set according to their default values).

%% file: supplementary/figures/gt/gt.tex
\begin{figure}
    \centering
    \includegraphics[width=0.48\textwidth]{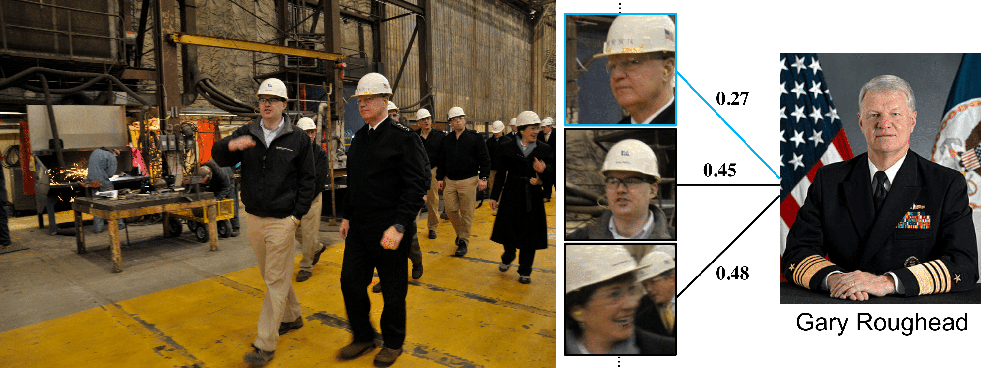}
    \caption{For each referred person associated with a ``primary" image on Wikimedia Commons (right), we compute face dissimilarities between the face in the ``primary" image and all detected faces. By finding a minimum weight bipartite matching (over all referred people), we recover a partial matching from referred people to detections (for simplicity, we only show these dissimilarities for a single referred person and for a subset of faces in the image). The estimated link is shown in blue.}
    \label{fig:gt_supp}
\end{figure}

%% file: supplementary/03-baselines.tex
\section{Baselines} Next we provide more details on how we obtain the reported scores on the pretrained models we evaluate on the WikiPeople test set.

\medskip \noindent \textbf{Gupta \etal~\cite{gupta2020contrastive}.}
We download their two pretrained models, trained on either COCO~\cite{lin2014microsoft} or Flickr30 Entities~\cite{plummer2015flickr30k}, from their official code repository\footnote{\url{https://github.com/BigRedT/info-ground}}. Following their implementation, visual features are extracted using the Bottom-Up Attention model ~\cite{anderson2018bottom} yielding a 2048-d visual representation. A pretrained BERT~\cite{devlin2018bert} model is used to extract 768-d contextualized word representations. We follow their evaluation protocol and compute a phrase-level attention score for each box by taking the maximum attention score assigned to the box by any of the tokens in
the name. The boxes are then ranked according to this phrase level score, with the maximum scoring box selected as the corresponding box. This top-scoring box is compared with the ground-truth box.

\medskip \noindent \textbf{SL-CCRF~\cite{liu2019learning}}. 
We download the pretrained ``Soft-Label Chain CRF Model'' from their official code repository\footnote{\url{https://github.com/liujch1998/SoftLabelCCRF}}, which yields the highest performance among their available models. Following their implementation, visual features are extracted using the Bottom-Up Attention model ~\cite{anderson2018bottom} yielding a 2048-d visual representation. 
We use their all default parameters, as follows: 1024-d contextualized word embeddings, the maximum number of mentions is set to 25, and a 5-d spatial feature is concatenated with the visual features.
The number of regions proposals are according to the number of detected people boxes.
However, as their model also includes a regression bounding box loss, their final predictions aren't entirely aligned with the input bounding boxes. We account for that gap in the evaluation, by considering boxes with IoU $\geq 0.5$.

\medskip \noindent \textbf{MAttNet~\cite{yu2018mattnet}.}
We downloaded a model from the official repository\footnote{\url{https://github.com/lichengunc/MAttNet}} that was pretrained on the RefCOCOg dataset~\cite{Mao2016GenerationAC}. Following their implementation, visual features are extracted using a modified implementation of Mask R-CNN~\cite{He2020MaskR}, as specified by the authors~\cite{yu2018mattnet}. However, we provide our own bounding boxes and compute Faster R-CNN region features~\cite{Ren2015FasterRT} over these, instead of using their proposals. A Language Attention Network with bi-directional LSTMs (as specified by MAttNet~\cite{yu2018mattnet}) is used to extract phrase embeddings. We use these modules to predict a detection for each individual referring expression (\ie a person's name).

%% file: supplementary/04-additional-results.tex
\input{supplementary/tables/baselines}
\input{supplementary/tables/ablations}

\input{supplementary/figures/analysis/analysis}

\section{Additional Results and Ablations}
We report performance obtained on all three baselines while training on our data in Table \ref{tab:baselines}. The low performance obtained on the baselines is not surprising as (1) weakly supervised techniques (such as Gupta et al.~\cite{gupta2020contrastive}) do not have access to ground truth supervision---in our ablations this similarly results in a significant performance drop; (2) phrase grounding techniques (such as MAttNet~\cite{yu2018mattnet}) only process the phrase describing the region (which would be masked out in our case); and (3) SL-CCRF also processes the masked out phrases, along with dependencies between string-adjacent phrases (which evidently are not enough on their own for the model to learn meaningful grounding).

All results reported in the paper are obtained by selecting, for each referred person, the most similar box according to $S$. In Table \ref{tab:ablations-supp}, we also report performance by performing a minimum weight bipartite matching~\cite{Kuhn1955TheHM} over the similarity matrix, thus producing a natural one-to-one mapping. As illustrated in the table, this yields a decrease in performance of approximately $1\%$. We also train a model with an additional (unsupervised) optimal transport loss, which was proposed for pretraining the UNITER~\cite{chen2020uniter} model, as it encourages sparsity, and could potentially improve alignments between words and regions in the image (or names and people's boxes in our case). Results show that adding this loss on top of $S$ does not yield an improvement in performance (and even slightly degrades our full model's performance). This suggests that robust alignments are achieved from the training supervision directly, without need for additional regularization.

Figure~\ref{fig:analysis} illustrates the distribution of samples and performance breakdowns for L$\rightarrow$R (Largest) and our model over the numbers of referred people in a caption ($n$) and people detected in an image ($m$). We compute average accuracies over all relevant test subset images. As illustrated in the figure, the heuristic surpasses our model over only two subsets---$(m = 2, n = 1)$ and $(m \ge 4, n \ge 4)$, given $m$ detections and $n$ referred people---and performs worse in all other subsets.

We find that occupations correlate with different situations---images featuring athletes, for instance, have different properties from those featuring singers. We observe that model performance varies somewhat across different occupation types. For instance, considering only the interactive subset of test samples, %
accuracy on people with athletic occupations (association football player, basketball player, etc.)\ is lower than accuracy for politicians or performers (actor, model, musician, etc.), while their distribution in the training set is similar (athletes, politicians, and performers are each captured by $10$--$13\%$ of the interactive training set). A potential explanation is that interactions within sports-themed images are broader and more complex than in other categories. %

We also observe that over the full set test, performance over politician samples is significantly lower, and this is also reflected in a lower left-to-right ordering accuracy. A visual analysis reveals that these samples are indeed more challenging, as in many cases the captions mostly mention notable individuals regardless of the visual arrangement of the captured individuals.

\input{supplementary/tables/occupation}

Finally, we experiment with training models using several forms of standard augmentation techniques. Results are reported in Table~\ref{tab:augmentations}. Note that the nature of our dataset and task renders some augmentations more sensible than others. In particular, a model trained with random horizontal flipping yields significantly lower performance. This is likely due to the inherent left-to-right ordering in the images and captions, as some captions in our dataset either explicit annotate people with ``(left)'' and ``(right)'', or implicitly mention people in the left-to-right order they appear in the image. %
Other augmentations, such as translating all bounding boxes within the image or performing random color jittering on the images, yields comparable performance. %
\input{supplementary/tables/augmentations}

\input{supplementary/figures/examples/results_supp}

\input{supplementary/figures/baselines/baselines}

\input{supplementary/figures/baselines/gupta}

\section{Additional Qualitative Results}
Figure~\ref{fig:results2-2} shows additional visualizations of our model's predictions for samples in our test set.
Figure~\ref{fig:baselines} and Figure~\ref{fig:gupta} respectively show results obtained with prior supervised and weakly-supervised grounding models.
As illustrated in the figures, prior visual grounding works struggle in correctly linking people across images and text for these challenging examples, which cover various interactions between multiple people. Errors can be attributed to selecting a single box for all referred people, or selecting (smaller) boxes that are unreferenced to in the caption.

%% file: supplementary/tables/baselines.tex
\begin{table}[t]
\centering
\setlength{\tabcolsep}{3.1pt}
\def\arraystretch{0.95}
\begin{tabularx}{0.51\columnwidth}{lccc}
\toprule
  Method  && Accuracy        \\ \midrule
  Gupta et al.~\cite{gupta2020contrastive} && 31.78 \\
    SL-CCRF~\cite{liu2020phrase} && 30.07 \\
      MAttNet~\cite{yu2018mattnet} && 27.53 \\
\bottomrule
\end{tabularx}
\vspace{3pt}
\caption{Performance obtained on the baselines trained on our data. As further detailed in the text, these baselines cannot be naively adapted for our task.}\label{tab:baselines}
\end{table}

%% file: supplementary/tables/ablations.tex
\begin{table}[t]
\centering
\setlength{\tabcolsep}{3.1pt}
\def\arraystretch{0.95}
\begin{tabularx}{0.87\columnwidth}{lccc}
\toprule
  Method  && Max-Box   & Bipartite     \\ \midrule
  \textbf{Input features} \\
 \quad w/o visual features && 55.4 & 54.2 \\
 \quad w/o spatial features &&  58.0 & 57.0 \\
 \quad w/o textual features && 51.3 & 50.8 \\ \midrule
 \textbf{Learning} \\
 \quad w/o $\mathcal{L}_\mathsf{intra}$ && 31.4 & 30.1 \\
  \quad  w/o $\mathcal{L}_\mathsf{inter}$ &&  61.9 & 60.9 \\
    \quad  w/o $\mathcal{L}_{\varnothing}$ && 61.7 &  61.2 \\
 \quad w/o pretraining &&  50.2 & 38.7 \\
\quad  w/ optimal transport loss && 62.2 & 61.6 \\
  \midrule
  \textbf{Ours (full)} &&  63.5 & 61.9  \\
\bottomrule
\end{tabularx}
\vspace{3pt}
\caption{Ablation study, evaluating the effect of using a bipartite
matching algorithm during inference (second column) and using an additional optimal transport loss (second to last row). }\label{tab:ablations-supp}
\end{table}

%% file: supplementary/figures/analysis/analysis.tex
\begin{figure*}
    \centering
\jsubfig{\includegraphics[height=5.0cm,trim={0 0.2cm 0 1.7cm},clip]{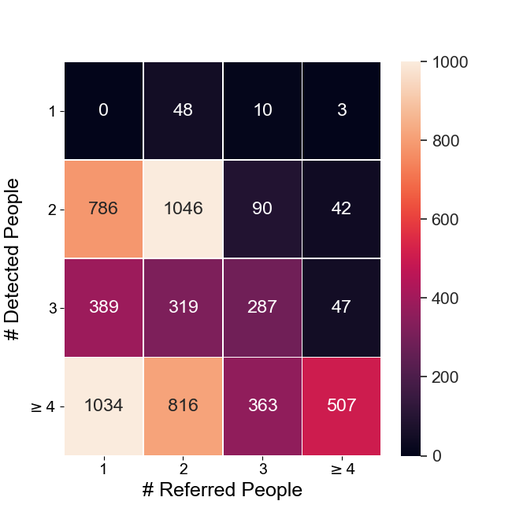}}{\vspace{-4pt}Distribution of Samples} 
\jsubfig{\includegraphics[height=5.0cm,trim={0 0.2cm 0 1.7cm},clip]{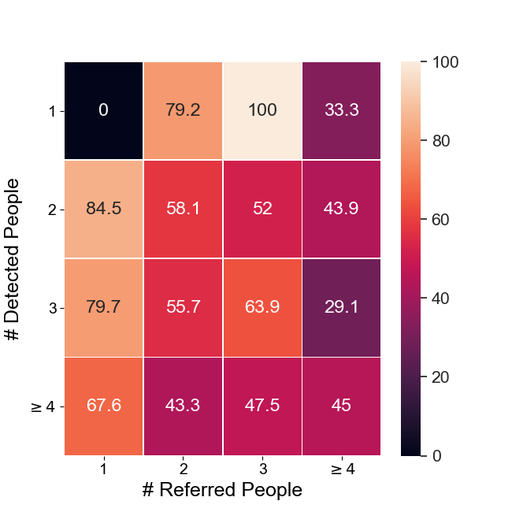}}{\vspace{-4pt}L$\rightarrow$R (Largest)} 
\jsubfig{\includegraphics[height=5.0cm,trim={0 0.2cm 0 1.7cm},clip]{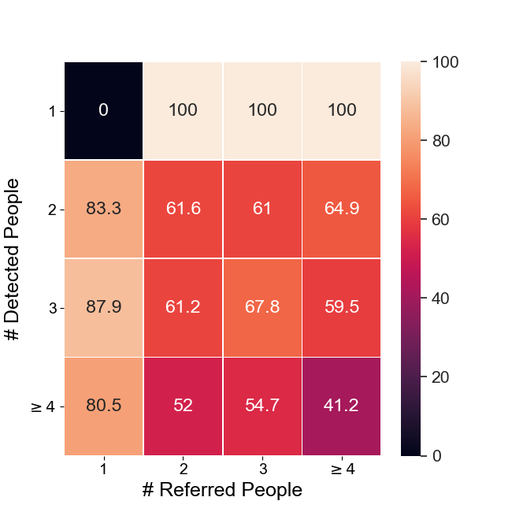}}{\vspace{-4pt}Ours}
     \vspace{5pt}
     \caption{Accuracy breakdown by number of referred people and detected faces for our model and the L$\rightarrow$R (Largest) baseline. The test set sample distribution is illustrated on the left (no images with just a single detection and referred person are included in our evaluation).}
    \label{fig:analysis}
\end{figure*}

%% file: supplementary/tables/occupation.tex
\begin{table}[t]
\centering
\setlength{\tabcolsep}{3.5pt}
\def\arraystretch{1.05}
\begin{tabularx}{0.94\columnwidth}{lcccccc}
\toprule
  Set  && Politicians & Athletes & Performers        \\ \midrule
  \textbf{Interactive}\\
  \quad L$\rightarrow$R (Largest)  && 47.1 & 43.0 & 49.8 \\
  \quad Ours && 52.5  & 51.1  & 54.9  \\
  \midrule
  \textbf{All} \\
  \quad L$\rightarrow$R (Largest)  && 52.4 & 70.6 & 67.4 \\
  \quad Ours && 54.8  & 76.3  & 71.2  \\
\bottomrule
\end{tabularx}
\vspace{3pt}
\caption{Analyzing model performance by identity occupation for the interactive subset and for all data samples. Test accuracy for the strongest baseline and for our model is reported for samples belonging to the occupation categories specified on top.}\label{tab:occupation}
\end{table}

%% file: supplementary/tables/augmentations.tex
\begin{table}[t]
\centering
\def\arraystretch{0.95}
\begin{tabularx}{0.655\columnwidth}{lccc}
\toprule
  Augmentation  && Accuracy        \\ \midrule
 \quad Ours && 63.5 \\
 \quad w/ horizontal flips &&  53.8  \\
 \quad w/ translations && 62.0 \\
  \quad w/ color jittering && 63.0 \\

\bottomrule
\end{tabularx}
\vspace{3pt}
\caption{Evaluating the effect of using standard data augmentation techniques during training.}\label{tab:augmentations}
\end{table}

%% file: supplementary/figures/examples/results_supp.tex
\begin{figure*} %
\jsubfig{\includegraphics[height=4.7cm]{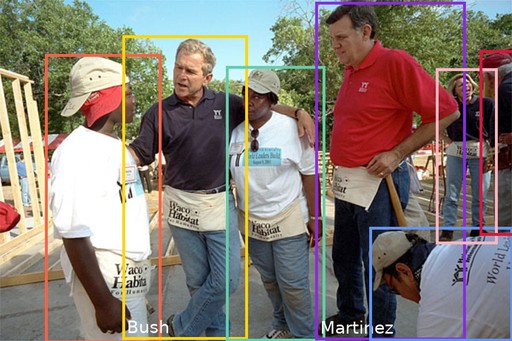}}{\vspace{-8pt} \begin{flushleft}
President \colorbox{gold}{\textbf{Bush}} and Secretary for Housing and Urban Development \textbf{Martinez}, far right, talk with new friends during a break from their house-building efforts at the Waco, Texas, location of Habitat for Humanity's "World Leaders Build" construction drive August 8, 2001.\end{flushleft}} \hfill
\jsubfig{\includegraphics[height=4.7cm]{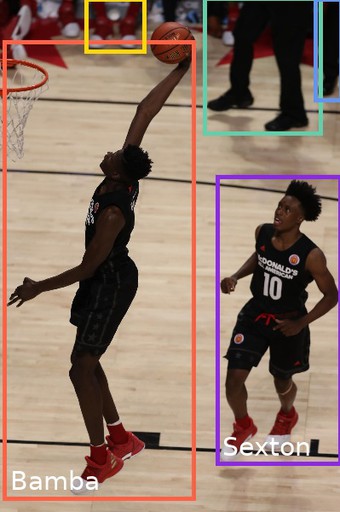}}{\vspace{-8pt}\begin{flushleft} \colorbox{tomato}{\textbf{Mohamed Bamba}} dunks in front of \textbf{Collin Sexton} at the McDonald's All-American Boys Game. \end{flushleft}} \hfill %
\jsubfig{\includegraphics[height=4.7cm,clip]{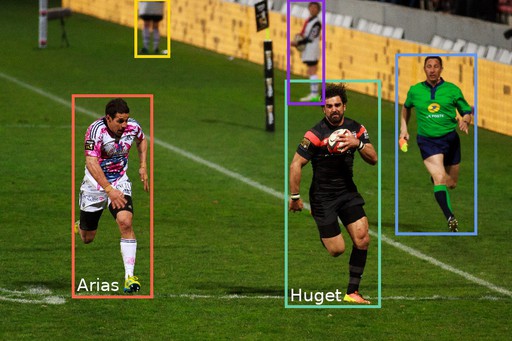}}{\vspace{-8pt}\begin{flushleft}\colorbox{mediumaquamarine}{\textbf{Yoann Huget}} out run \textbf{Julien Arias} to score his second try of the match during Stade toulousain vs Stade français Paris, March 24th, 2012. \end{flushleft}} \\
\jsubfig{\includegraphics[height=4.8cm,clip]{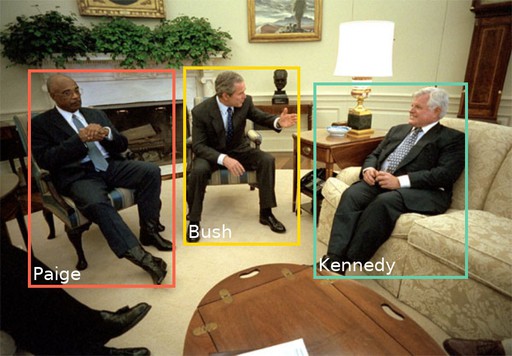}}{\vspace{-8pt}\begin{flushleft}
President \colorbox{gold}{\textbf{Bush}} meets with Secretary of Education \colorbox{tomato}{\textbf{Rod Paige}}, left, and Senator \colorbox{mediumaquamarine}{\textbf{Edward Kennedy}} August 2, 2001, to discuss the education reforms for the country.\end{flushleft}}\hfill
\jsubfig{\includegraphics[height=4.8cm]{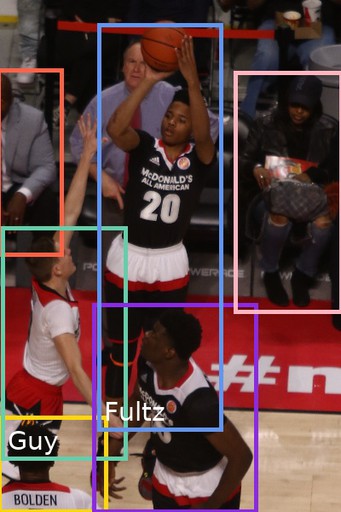}}{\vspace{-8pt} \begin{flushleft}\colorbox{cornflowerblue}{\textbf{Markelle Fultz}} shoots over \textbf{Kyle Guy} at the McDonald's All-American Boys Game.  \end{flushleft}}\hfill %
\jsubfig{\includegraphics[height=4.8cm, clip]{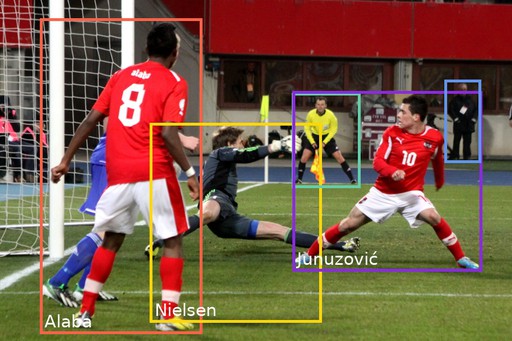}}{\vspace{-8pt}\begin{flushleft}The photo shows \textbf{David Alaba} (Austria), \colorbox{gold}{\textbf{Gunnar Nielsen}} (Faroe Islands) \colorbox{blueviolet}{\textbf{Zlatko Junuzović}} (Austria). \end{flushleft}} \\
\jsubfig{\includegraphics[height=5.7cm]{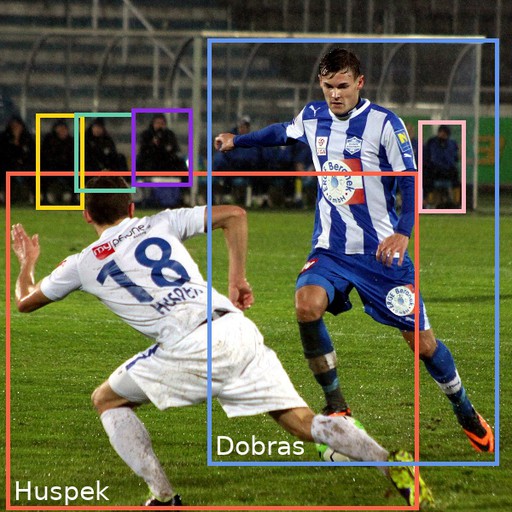}}{\vspace{-8pt} \begin{flushleft} The photo shows \colorbox{cornflowerblue}{\textbf{Kristijan Dobras}} (SC Wiener Neustadt, blue shirt) and Philipp Huspek (SV Grödig, white shirt). \end{flushleft}}\hfill
\jsubfig{\includegraphics[height=5.7cm]{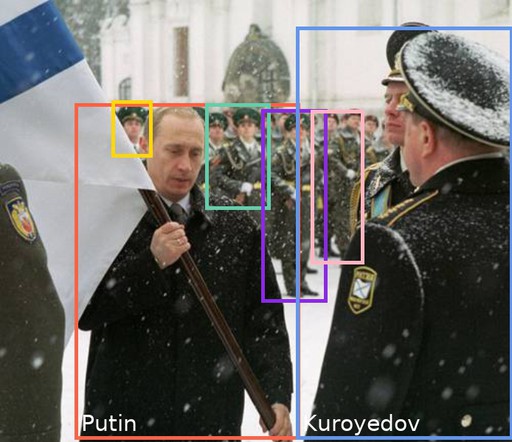}}{\vspace{-8pt}\begin{flushleft}
President \colorbox{tomato}{\textbf{Putin}} presenting the banner of the Navy to its Commander-in-Chief Admiral Vladimir Kuroyedov.\end{flushleft}}\hfill
\jsubfig{\includegraphics[height=5.7cm]{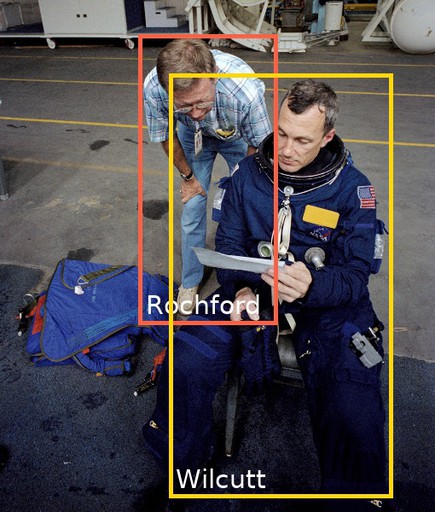}}{\vspace{-8pt} \begin{flushleft}Astronaut \colorbox{gold}{\textbf{Terrence W. Wilcutt}}, STS-68 pilot, goes over his notes. Checking the notes is \textbf{Alan M. Rochford}, suit expert.
\end{flushleft}}
\vspace{-7pt}
\caption{\textbf{Additional box--name correspondences predicted by our model.} We show predicted entities on top of the their associated box (in white). Ground truth links are denoted by matching colors. }
\label{fig:results2-2}
\end{figure*}

%% file: supplementary/figures/baselines/baselines.tex
\begin{figure*} %
\centering
\rotatebox{90}{SL-CCRF \cite{liu2020phrase}}
\jsubfig{\includegraphics[height=4.4cm]{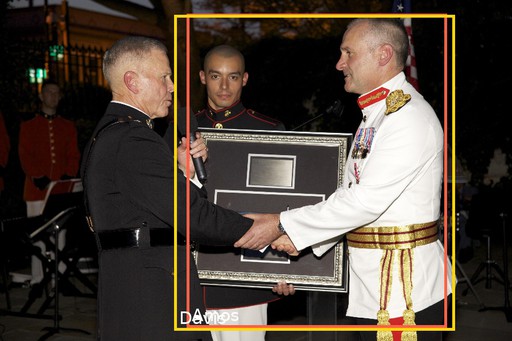}}{}\hfill
\jsubfig{\includegraphics[height=4.4cm]{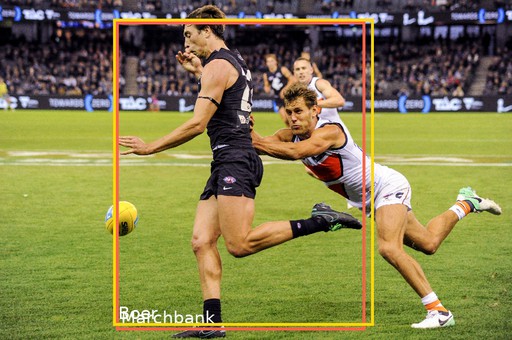}}{} \hfill
\jsubfig{\includegraphics[height=4.4cm]{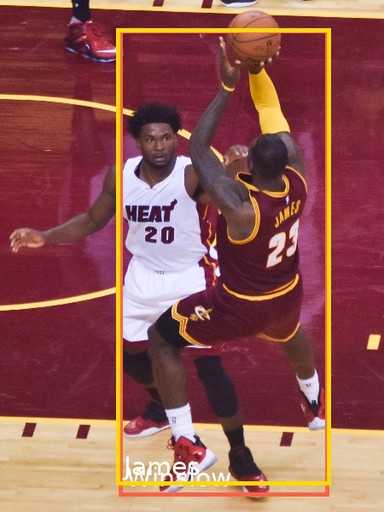}}{} \vspace{2pt} \\
\rotatebox{90}{MAttNet \cite{yu2018mattnet} }
\jsubfig{\includegraphics[height=4.4cm]{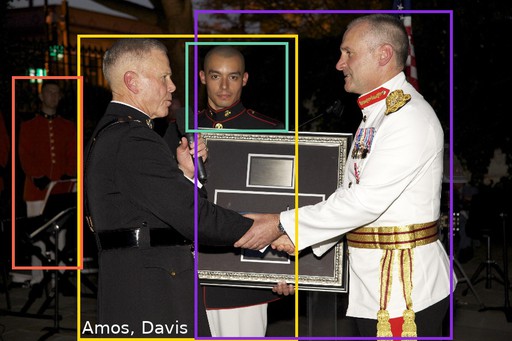}}{}\hfill
\jsubfig{\includegraphics[height=4.4cm]{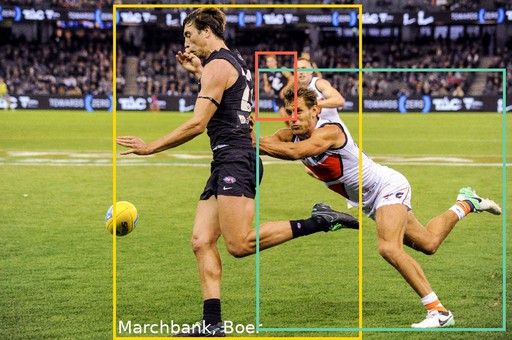}}{} \hfill
\jsubfig{\includegraphics[height=4.4cm]{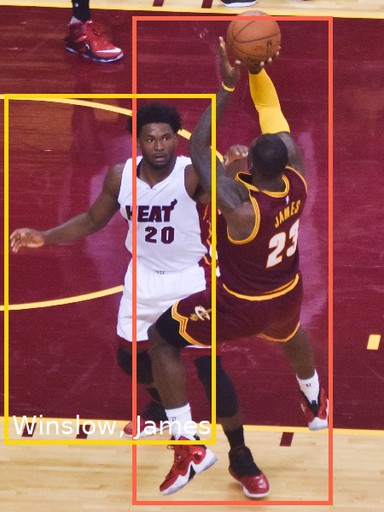}}{}
\\ 
\rotatebox{90}{UNITER \cite{chen2020uniter} }
\jsubfig{\includegraphics[height=4.4cm]{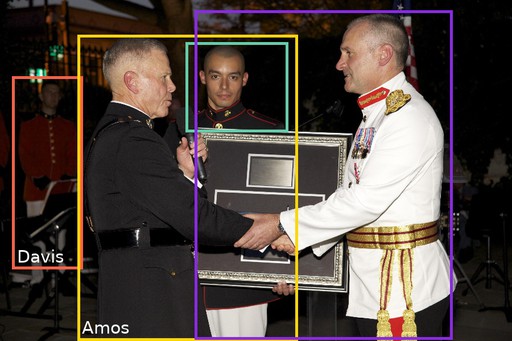}}{}\hfill
\jsubfig{\includegraphics[height=4.4cm]{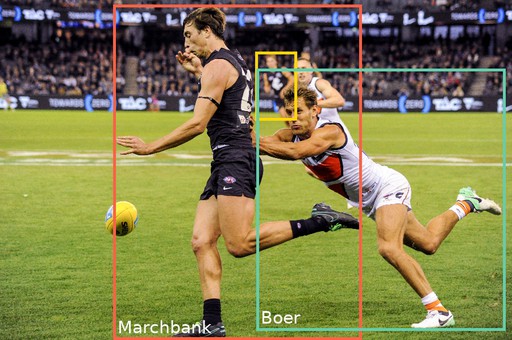}}{} \hfill
\jsubfig{\includegraphics[height=4.4cm]{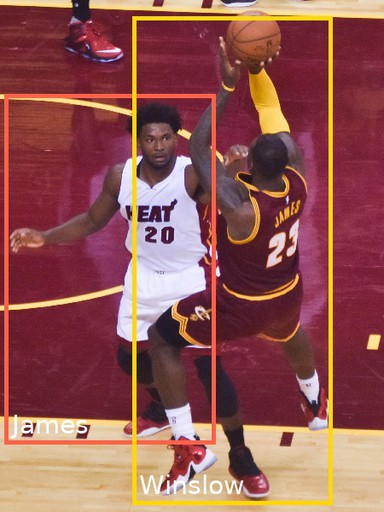}}{}
\vspace{2pt} \\
\rotatebox{90}{Ours }
\jsubfig{\includegraphics[height=4.4cm]{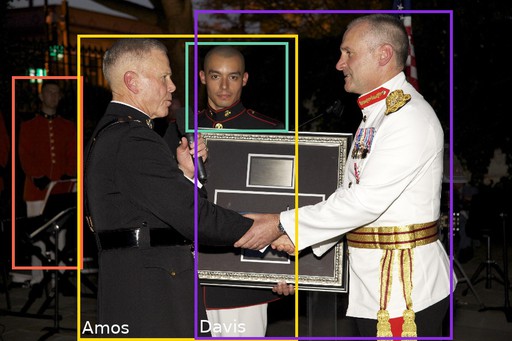}}{\vspace{-8pt}\begin{flushleft}
Commandant of the U.S. Marine Corps Gen. \colorbox{gold}{\textbf{James F. Amos}}, left, participates in a gift exchange with Commandant General of the British Royal Marines Maj. Gen. \textbf{Ed Davis}. %
\end{flushleft}}\hfill
\jsubfig{\includegraphics[height=4.4cm]{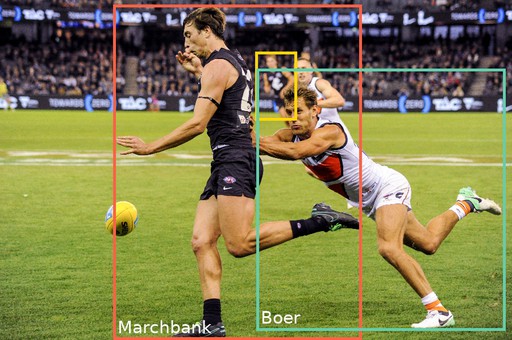}}{\vspace{-8pt}{\begin{flushleft} \colorbox{tomato}{\textbf{Caleb Marchbank}} kicking away from \textbf{Matt de Boer} during the AFL round twelve match between Carlton and Greater Western Sydney on 11 June 2017 at Etihad Stadium. %
\end{flushleft}}} \hfill
\jsubfig{\includegraphics[height=4.4cm]{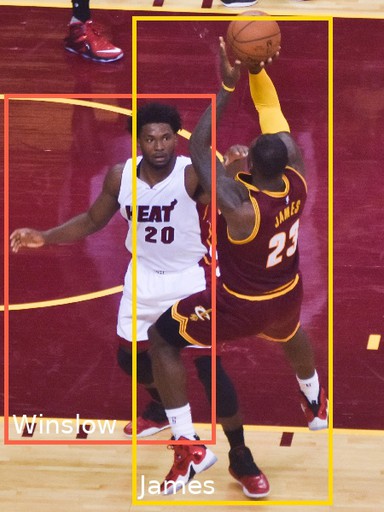}}{\vspace{-8pt} \begin{flushleft}\colorbox{tomato}{\textbf{Justise Winslow}} of the Miami Heat defending \textbf{LeBron James}. %
\end{flushleft}}
\vspace{-7pt}
\caption{Comparing against supervised visual grounding techniques, SL-CCRF~\cite{liu2020phrase} and MAttNet~\cite{yu2018mattnet}, and the pretrained UNITER~\cite{chen2020uniter} model. We show predicted entities on top of the their associated box (in white). Ground truth links are denoted by matching colors. For SL-CCRF~\cite{liu2020phrase}, as their model incorporates a regression loss that modifies the input boxes, we only show the predicted boxes. In both SL-CCRF~\cite{liu2020phrase} and MAttNet~\cite{yu2018mattnet}, errors are attributed to selecting the same box for multiple referred people. It should be noted that this is not always the case, and from further visual inspection, in many cases these models are capable of selecting multiple boxes. We can see that the pretrained UNITER model provides unique assignments for all three examples, possibly due to the optimal transport loss they propose to encourage robust word-region alignments. The selected boxes, however, are only accurate in the middle example (and partially accurate in the leftmost example). }
\label{fig:baselines}
\end{figure*}

%% file: supplementary/figures/baselines/gupta.tex
\begin{figure*} %
\centering
\rotatebox{90}{Gupta \etal (Flickr30K Entities)}
\jsubfig{\includegraphics[height=4.4cm]{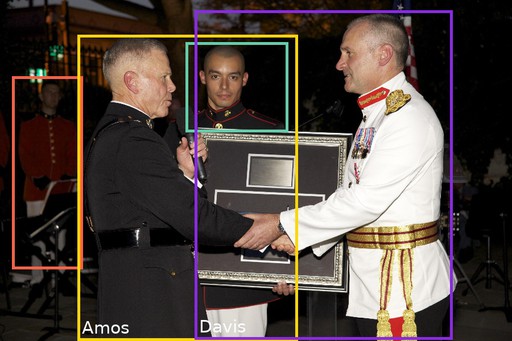}}{}\hfill
\jsubfig{\includegraphics[height=4.4cm]{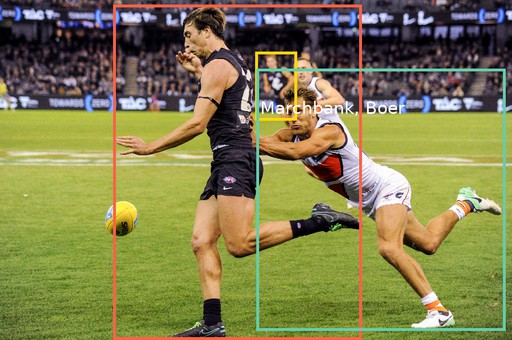}}{} \hfill
\jsubfig{\includegraphics[height=4.4cm]{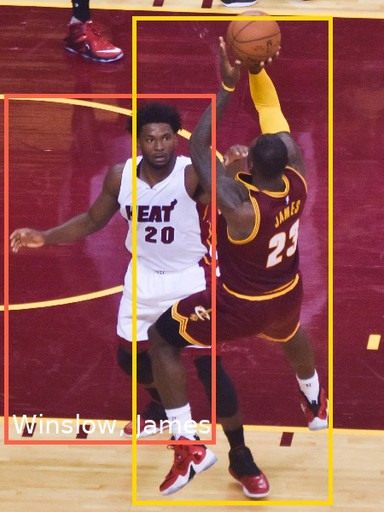}}{} \vspace{2pt} \\
\rotatebox{90}{Gupta \etal (COCO)}
\jsubfig{\includegraphics[height=4.4cm]{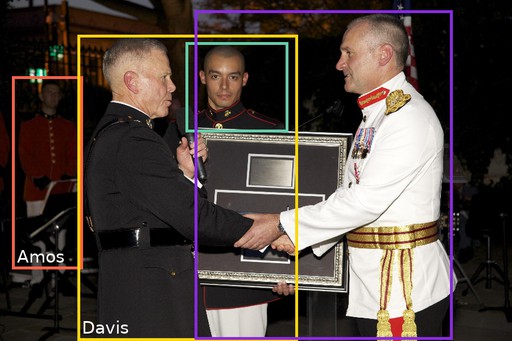}}{}\hfill
\jsubfig{\includegraphics[height=4.4cm]{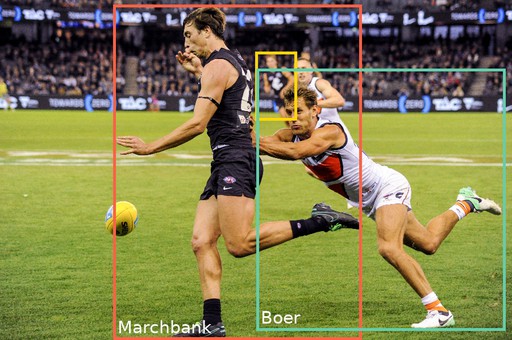}}{} \hfill
\jsubfig{\includegraphics[height=4.4cm]{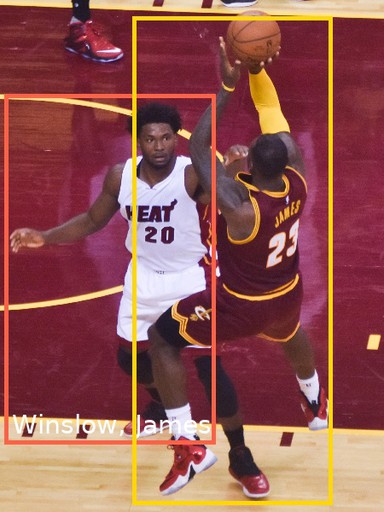}}{} \vspace{2pt}
\\ 
\rotatebox{90}{Ours }
\jsubfig{\includegraphics[height=4.4cm]{supplementary/figures/examples/090476_with_boxes.jpg}}{\vspace{-8pt}\begin{flushleft}
Commandant of the U.S. Marine Corps Gen. \colorbox{gold}{\textbf{James F. Amos}}, left, participates in a gift exchange with Commandant General of the British Royal Marines Maj. Gen. \textbf{Ed Davis}. %
\end{flushleft}}\hfill
\jsubfig{\includegraphics[height=4.4cm]{supplementary/figures/examples/050824_with_boxes.jpg}}{\vspace{-8pt}{\begin{flushleft} \colorbox{tomato}{\textbf{Caleb Marchbank}} kicking away from \textbf{Matt de Boer} during the AFL round twelve match between Carlton and Greater Western Sydney on 11 June 2017 at Etihad Stadium. %
\end{flushleft}}} \hfill
\jsubfig{\includegraphics[height=4.4cm]{supplementary/figures/examples/141640_with_boxes.jpg}}{\vspace{-8pt} \begin{flushleft}\colorbox{tomato}{\textbf{Justise Winslow}} of the Miami Heat defending \textbf{LeBron James}. %
\end{flushleft}}
\vspace{-7pt}
\caption{Comparing against the weakly-supervised visual grounding technique proposed by Gupta \etal \cite{gupta2020contrastive}. We evaluate on both of their pretrained models, trained on Flickr30K Entities~\cite{plummer2015flickr30k} (top row) and COCO~\cite{lin2014microsoft} (second row). We show predicted entities on top of the their associated box (in white). Ground truth links are denoted by matching colors. Errors are attributed to either selecting the same box for multiple referred people (\eg rightmost example), or selecting irrelevant boxes, such as the yellow box in the middle image, top row, or the orange box in the left image, second row.
 }
\label{fig:gupta}
\end{figure*}